\documentclass[twoside,11pt]{article}
\pdfoutput=1

\usepackage{jmlr2e}

\usepackage[letterpaper=true,pdfview=FitH,pdfstartview=FitH,bookmarks=true]{hyperref}

\usepackage{algorithm}
\usepackage{algorithmic}
\usepackage{amsfonts}
\usepackage{amsmath}
\usepackage{amssymb}
\usepackage{latexsym}
\usepackage{enumerate}
\usepackage{graphicx}
\usepackage{url}
\usepackage{lastpage}			

\jmlrheading{1}{2009}{1-\pageref{LastPage}}{11/09}{?}{Benjamin I. P. Rubinstein and J. Hyam Rubinstein}

\ShortHeadings{A Geometric Approach to Sample Compression}{Rubinstein and Rubinstein}
\firstpageno{1}

\newtheorem{question}[theorem]{Question}
\newcommand{\term}[1]{\emph{#1}}
\renewcommand{\vec}[1]{\mathbf{#1}}
\newcommand{\cC}{\mathcal{C}}
\renewcommand{\H}{\mathcal{H}}

\newcommand{\bbH}{\mathbb{H}}
\newcommand{\R}{\mathbb{R}}
\newcommand{\cN}{\mathcal{N}}
\newcommand{\N}{\mathbb{N}}

\newcommand{\cO}{\mathcal{O}}
\newcommand{\cP}{\mathcal{P}}
\newcommand{\bR}{\mathbb{R}}

\newcommand{\nth}[2]{{#1}^{\mbox{\scriptsize {#2}}}}
\newcommand{\Construct}[2]{\mbox{\sc{MaximumClasses}}(#1,#2)}

\newcommand{\smallchoose}[2]{\left(\begin{smallmatrix}{#1}\\{#2}\end{smallmatrix}\right)}
\newcommand{\bigchoose}[2]{{{#1}\choose{#2}}}
\newcommand{\Oinc}[1]{\mathcal{G}\!\left(#1\right)}
\newcommand{\incidentdims}[2]{I_{#1}\!(#2)}
\newcommand{\VC}[1]{\mathrm{VC}\!\left(#1\right)}

\newcommand{\proj}[2]{\Pi_{#1}\!\left(#2\right)}
\newcommand{\card}[1]{\left|#1\right|}

\newcommand{\tail}[2]{\mathrm{tail}_{#1}\left(#2\right)}

\begin{document}

\title{A Geometric Approach to Sample Compression}

\author{\name Benjamin I. P. Rubinstein \email benr@cs.berkeley.edu \\
       \addr Computer Science Division\\
       University of California\\
       Berkeley, CA 94720-1776, USA
       \AND
       \name J. Hyam Rubinstein \email rubin@ms.unimelb.edu.au \\
       \addr Department of Mathematics \& Statistics\\
       University of Melbourne\\
       Parkville, Victoria 3010, Australia}

\editor{Leslie Pack Kaelbling}

\maketitle

\begin{abstract}
The Sample Compression Conjecture of Littlestone \& Warmuth has remained
unsolved for over two decades. While maximum classes (concept classes meeting
Sauer's Lemma with equality) can be compressed, the compression of general
concept classes reduces to compressing maximal classes (classes that cannot be
expanded without increasing VC-dimension). Two promising ways forward are:
embedding maximal classes into maximum classes with at most a polynomial
increase to
VC dimension, and compression via operating on geometric representations. This
paper presents positive results on the latter approach and a first negative
result on the former, through a systematic investigation of finite maximum
classes.  Simple arrangements of hyperplanes in Hyperbolic space are shown to
represent maximum
classes, generalizing the corresponding Euclidean result.  We show that 
sweeping a generic hyperplane across such arrangements forms an unlabeled
compression scheme of size VC dimension and corresponds to a special case of
peeling the one-inclusion graph, resolving a recent conjecture of
Kuzmin \& Warmuth. A bijection between finite maximum classes and
certain arrangements of Piecewise-Linear (PL) hyperplanes in either a
ball or Euclidean space is established.  Finally we show that $d$-maximum 
classes corresponding to PL hyperplane arrangements in $\R^d$ have
cubical complexes homeomorphic to a $d$-ball, or equivalently
complexes that are manifolds with boundary.
A main result is that PL arrangements can be swept by a moving hyperplane to
unlabeled $d$-compress {\em any} finite maximum class, forming a peeling scheme
as conjectured by Kuzmin \& Warmuth. A corollary is that some $d$-maximal
classes cannot be embedded into any maximum class of VC dimension $d+k$, for any
constant $k$. The construction of the PL sweeping involves {\em Pachner moves}
on the one-inclusion graph, corresponding to moves of a hyperplane across the
intersection of $d$ other hyperplanes. This extends the well known Pachner moves
for triangulations to cubical complexes.
\end{abstract}


\section{Introduction}

\term{Maximum} concept classes have the largest cardinality possible for their
given VC dimension. Such classes are of particular interest as their
special recursive structure underlies all general sample compression
schemes known to-date~\citep{F89,W03,KW07}.  It is this structure
that admits many elegant geometric and algebraic topological representations
upon which this paper focuses.

\citet{LW86} introduced the study of
\term{sample compression schemes}, defined as a pair of mappings for
given concept class $C$: a \term{compression function} mapping a $C$-labeled
$n$-sample to a subsequence of labeled examples and a \term{reconstruction} \term{function} mapping the
subsequence to a concept consistent with the entire $n$-sample.  A compression
scheme of bounded size---the maximum cardinality of the subsequence image---was
shown to imply learnability.  The converse---that classes of VC
dimension $d$ admit compression schemes of size $d$---has become one of
the oldest unsolved problems actively pursued within learning theory~\citep{F89,HSW92,BDL98,W03,H06,KW07,RBR07a,RBR08,RR08}. Interest in the 
conjecture has been motivated by its interpretation as the converse to the
existence of compression bounds for PAC learnable classes~\citep{LW86}, the
basis of practical machine learning methods on
compression schemes~\citep{MS03,LBS04}, and
the conjecture's connection to a deeper
understanding of the combinatorial properties of concept
classes~\citep{RBR08,RR08}.
Recently~\citet{KW07} achieved compression of maximum classes without the use of
labels.  They also conjectured that their elegant
Min-Peeling
Algorithm constitutes such an unlabeled $d$-compression scheme for
$d$-maximum classes.

As discussed in our previous work~\citep{RBR08}, maximum classes can be 
fruitfully viewed as \term{cubical complexes}. These are also topological
spaces, with each cube equipped with a natural topology of open sets
from its standard embedding into Euclidean space. 
We proved that $d$-maximum classes correspond
to \term{$d$-contractible complexes}---topological spaces with an identity map
homotopic to a constant map---extending the result that $1$-maximum classes
have trees for one-inclusion graphs.  Peeling can be viewed as a
special form of contractibility for maximum classes. However, there are many 
non-maximum contractible cubical complexes that cannot be
peeled, which demonstrates that peelability reflects more detailed structure
of maximum classes than given by contractibility alone. 

In this paper we approach peeling from the direction of simple hyperplane
arrangement representations of maximum classes. \citet[Conjecture~1]{KW07}
predicted that $d$-maximum classes corresponding to simple linear
hyperplane arrangements could be unlabeled $d$-compressed by sweeping a
generic hyperplane across the arrangement, and that concepts are min-peeled as
their corresponding cell is swept away. We
positively resolve the first part of the conjecture and show that sweeping
such arrangements corresponds to a new form of \term{corner-peeling}, which we
prove is distinct from min-peeling. While \term{min-peeling} removes
minimum degree concepts from a one-inclusion graph,
corner-peeling peels vertices that are contained in unique cubes of maximum
dimension.  

We explore simple hyperplane arrangements in Hyperbolic geometry, which we
show correspond to a set of maximum classes, properly containing those
represented by simple linear Euclidean arrangements.  These classes can again
be corner-peeled by sweeping.  Citing the proof of existence of maximum
unlabeled compression schemes due to~\citet{BDL98}, \citet{KW07} ask
whether unlabeled compression schemes for infinite classes such as
positive half spaces can be constructed explicitly. We present
constructions for illustrative but simpler classes, suggesting that there are
many interesting infinite maximum classes admitting explicit compression
schemes, and under appropriate conditions, sweeping infinite Euclidean,
Hyperbolic or PL arrangements corresponds to compression by 
corner-peeling.

Next we prove that all maximum classes in $\{0,1\}^n$ are represented as
simple arrangements of Piecewise-Linear (PL) hyperplanes in the $n$-ball. 
This extends previous work by~\citet{GW94} on viewing simple PL hyperplane
arrangements as maximum classes.
The close relationship between
such arrangements and their Hyperbolic versions suggests that they could be
equivalent. Resolving the main problem left open in the preliminary version of
this paper,~\citep{RR08}, we show that sweeping of $d$-contractible PL
arrangements does compress all finite 
maximum classes by corner-peeling, completing~\citep[Conjecture~1]{KW07}.

We show that a one-inclusion graph $\Gamma$ can be represented by a
$d$-contractible PL hyperplane arrangement if and only if $\Gamma$ is a
strongly contractible cubical complex. 
This motivates the nomenclature of $d$-contractible for the class of arrangements of PL hyperplanes.
Note then that these one-inclusion graphs admit a corner-peeling scheme of the same 
size $d$ as the largest dimension of a cube in $\Gamma$. Moreover if such a graph $\Gamma$ admits a corner-peeling scheme, then it is a contractible cubical complex. We give a simple example to show that there are one-inclusion graphs which admit corner-peeling schemes but are not strongly contractible and so are not represented by a $d$-contractible PL hyperplane arrangement. 

Compressing \term{maximal classes}---classes which cannot be grown without an
increase to their VC dimension---is sufficient for compressing all classes,
as embedded classes trivially inherit compression schemes of
their super-classes.  This reasoning motivates the attempt to
embed $d$-maximal classes into $O(d)$-maximum
classes~\citep[Open Problem~3]{KW07}.  We present non-embeddability results
following from our earlier counter-examples to Kuzmin \& Warmuth's minimum 
degree conjecture~\citep{RBR08}, and our new results on corner-peeling.
We explore with examples, maximal classes that can be compressed but
not peeled, and classes that are not strongly contractible but can be
compressed.  

Finally, we investigate algebraic topological properties of maximum classes. Most notably
we characterize $d$-maximum classes, corresponding to simple linear
Euclidean arrangements, as cubical complexes homeomorphic to the $d$-ball. 
The result that such classes' boundaries are homeomorphic to the $(d-1)$-sphere
begins the study of the boundaries of maximum classes, which are closely
related to peeling. We conclude with several open problems.

\section{Background}

\subsection{Algebraic Topology}

\begin{definition}
A \term{homeomorphism} is a one-to-one and onto map $f$ between topological spaces
such that both $f$ and $f^{-1}$ are continuous.  Spaces $X$ and $Y$ are said
to be \term{homeomorphic} if there exists a homeomorphism $f: X\to Y$.
\end{definition}

\begin{definition}
A \term{homotopy} is a continuous map $F:X \times [0,1]\to Y$. The \term{initial
map} is $F$ restricted to $X\times \{0\}$ and the \term{final map} is $F$
restricted to $X\times\{1\}$. We say that the initial and final maps are
\term{homotopic}.
A \term{homotopy equivalence} between spaces $X$ and $Y$ is a pair of maps
$f: X\to Y$ and $g: Y\to X$ such that $f\circ g$ and  $g\circ f$ are homotopic to the identity maps on $Y$
and $X$ respectively.  We say that
$X$ and $Y$ have the \term{same homotopy type} if there is a homotopy equivalence between them. 
A \term{deformation retraction} is a special homotopy equivalence between a space $X$ and a subspace $A \subseteq X$.
It is a continuous map $r:X \to X$ with the properties that the restriction of $r$ to $A$ is the identity map on $A$, $r$ has range $A$ and $r$ is
homotopic to the identity map on $X$.
\end{definition}

\begin{definition}
A \term{cubical complex} is a union of solid cubes of the form
$[a_1,b_1]\times\ldots\times[a_m,b_m]$, for bounded $m\in\N$,
such that the intersection of any two cubes in the complex is either a
cubical face of both cubes or the empty-set.
\end{definition}

\begin{definition}
A \term{contractible cubical complex} $X$ is one which has the same homotopy type as
a one point space $\{p\}$. $X$ is contractible if and only if the constant map from $X$ to $p$ 
is a homotopy equivalence. 
\end{definition}

\subsection{Concept Classes and their Learnability}

A \term{concept class} $C$ on \term{domain} $X$, is a subset of the power set
of set $X$ or equivalently $C\subseteq\{0,1\}^X$.  We primarily consider
finite domains and so will write $C\subseteq\{0,1\}^n$ in the sequel, where
it is understood that $n=\card{X}$ and the $n$ dimensions or \term{colors} are
identified with an ordering $\{x_i\}_{i=1}^n=X$.

The \term{one-inclusion graph} $\Oinc{C}$ of $C\subseteq\{0,1\}^n$ is
the graph with vertex-set $C$ and edge-set containing $\{u,v\}\subseteq C$
iff $u$ and $v$ differ on exactly one component~\citep{HLW94}; $\Oinc{C}$ forms
the basis of a prediction strategy with essentially-optimal worst-case expected
risk.  $\Oinc{C}$ can be viewed as a simplicial complex in $\R^n$ by filling
in each face with a product of continuous
intervals~\citep{RBR08}. Each edge $\{u,v\}$ in $\Oinc{C}$ is labeled by the 
component on which the two vertices $u,v$ differ. 

A \term{$d$-complete collection} is a union of $d$-subcubes in $\{0,1\}^n$, one with each choice of $d$ colors from $n$.

Probably Approximately Correct learnability of a concept class
$C\subseteq\{0,1\}^X$ is characterized by the finiteness of
the Vapnik-Chervonenkis (VC) dimension of $C$~\citep{BEHW89}.  One key
to all such results is Sauer's Lemma.

\begin{definition}\label{def:vcdim}
The \term{VC-dimension} of concept class $C\subseteq\{0,1\}^X$ is defined as 
$\VC{C}=\sup\left\{n\,\left|\,\exists Y\in {X \choose n}, \proj{Y}{C}=\{0,1\}^n\right.\right\}$ 
where $\proj{Y}{C}=\left\{\left(c(x_1),\ldots,c(x_n)\right)\mid c\in C\right\}\subseteq\{0,1\}^n$
is the \term{projection} of $C$ on sequence $Y=(x_1,\ldots,x_n)$.
\end{definition}

\begin{lemma}[\citealp{VC71,S72,SS72}]\label{lemma:sauer} The cardinality of any
concept classes $C\subseteq\{0,1\}^n$ is bounded by
$\card{C}\leq\sum_{i=1}^{\VC{C}}\!\bigchoose{n}{i}$.
\end{lemma}

Motivated by maximizing concept class cardinality under a fixed VC-dimension,
which is related to constructing general sample compression schemes (see
Section~\ref{sec:compression}), \citet{W87}~defined the following special 
classes.

\begin{definition}
Concept class $C\subseteq\{0,1\}^X$ is called \term{maximal} if
$\VC{C\cup\{c\}}>\VC{C}$ for all $c\in\{0,1\}^X\backslash C$.  Furthermore
if $\proj{Y}{C}$ satisfies Sauer's Lemma with equality for each
$Y\in{X \choose n}$, for every $n \in \N$, then $C$ is termed \term{maximum}.  If
$C\subseteq\{0,1\}^n$ then $C$ is maximum (and hence maximal) if $C$ meets
Sauer's Lemma with equality.
\end{definition}

The \term{reduction} of $C\subseteq\{0,1\}^n$ with respect to
$i\in[n]=\{1,\ldots,n\}$ is the class
$C^i=\proj{[n]\backslash\{i\}}{\left\{c\in C\mid i\in \incidentdims{\Oinc{C}}{c}\right\}}$ where $\incidentdims{\Oinc{C}}{c}\subseteq[n]$ denotes the 
labels of the edges incident to vertex $c$; a \term{multiple reduction} is the
result of performing several reductions in sequence.
The \term{tail} of class $C$ is
$\tail{i}{C}=\left\{c\in C\mid i\notin\incidentdims{\Oinc{C}}{c}\right\}$.
Welzl showed that if $C$ is $d$-maximum, then $\proj{[n]\backslash\{i\}}{C}$
and $C^i$ are maximum of VC-dimensions $d$ and $d-1$ respectively. 

The results presented below relate to other geometric and topological
representations of maximum classes existing in the literature. Under
the guise of `forbidden labels', \citet{F89} showed that maximum
$C\subseteq\{0,1\}^n$ of VC-dim $d$ is the union of a maximally overlapping
\term{$d$-complete collection of cubes}~\citep{RBR08}---defined as a collection
of $d$-cubes which project onto all $\smallchoose{n}{d}$ possible sets of $d$
coordinate directions. (An alternative proof was developed by \citealt{TN06}.)
It has long been known that VC-$1$ maximum classes
have one-inclusion graphs that are trees~\citep{D85}; we previously
extended this result by showing that when viewed as complexes, $d$-maximum
classes are contractible $d$-cubical complexes~\citep{RBR08}.  Finally the
cells of a simple linear arrangement of $n$ hyperplanes in $\R^d$ form a VC-$d$
maximum class in the $n$-cube~\citep{E87}, but not all finite maximum
classes correspond to such Euclidean arrangements~\citep{F89}.

\subsection{Sample Compression Schemes}
\label{sec:compression}

\citet{LW86} showed that the existence of a compression scheme of
finite size is sufficient for learnability of $C$, and conjectured the converse,
that $\VC{C}=d<\infty$ implies a compression scheme of size $d$. 
Later \citet{W03} weakened the conjectured size to $O(d)$.  To-date it is
only known that maximum classes can be $d$-compressed~\citep{F89}.  Unlabeled
compression was first explored by~\citet{BDL98}; \citet{KW07} defined unlabeled
compression as follows, and explicitly constructed schemes of size $d$ for
maximum classes.

\begin{definition}
Let $C$ be a $d$-maximum class on a finite domain $X$. 
A \term{representation mapping} $r$ of $C$ satisfies:
\begin{enumerate}
\item $r$ is a bijection between $C$ and subsets of $X$ of size at most $d$; and
\item $\mbox{[\emph{non-clashing}]}:$\ \ $c|\left(r(c)\cup r(c')\right)\neq c'|\left(r(c)\cup r(c')\right)$ for all $c,c'\in C$, $c\neq c'$.
\end{enumerate}
\end{definition}

As with all previously published labeled schemes, all previously known unlabeled
compression schemes for maximum classes exploit their special recursive
projection-reduction structure and so it is doubtful that such schemes will
generalize. 
\citet[Conjecture~2]{KW07} conjectured that their \term{Min-Peeling} Algorithm
constitutes an unlabeled $d$-compression scheme for maximum classes; it
iteratively removes minimum degree vertices from $\Oinc{C}$, representing
the corresponding concepts by the remaining incident dimensions in the
graph.  The authors also conjectured that sweeping
a hyperplane in general position across a simple linear arrangement forms
a compression scheme that corresponds to min-peeling the associated maximum
class~\citep[Conjecture~1]{KW07}.
A particularly promising approach to compressing general classes is via their
maximum-embeddings: a class $C$ embedded in class $C'$ trivially inherits any
compression scheme for $C'$, and so an important open problem is to embed
maximal classes into maximum classes with at most a linear increase in
VC-dimension~\citep[Open Problem~3]{KW07}.

\section{Preliminaries}

\subsection{Constructing All Maximum Classes} 
\label{sec:lifting}

The aim in this section is to describe an algorithm for constructing all
maximum classes of VC dimension $d$ in the $n$-cube. This process can be
viewed as the inverse of mapping a maximum class to its $d$-maximum projection
on $[n]\backslash\{i\}$ and the corresponding $(d-1)$-maximum reduction.

\begin{definition}\label{def:connected}
Let $C,C'\subseteq\{0,1\}^n$ be maximum classes of VC-dimensions $d,d-1$
respectively, so that $C' \subset C$, and let $C_1,C_2\subset C$ be $d$-cubes, i.e., $d$-faces of 
the $n$-cube $\{0,1\}^n$.
\begin{enumerate}[1.]
\item $C_1,C_2$ are \term{connected} if there exists a path in the one-inclusion graph $\Oinc{C}$ with
end-points in $C_1$ and $C_2$; and
\item $C_1,C_2$ are said to be \term{$C'$-connected} if there exists such a
connecting path that further does not intersect $C'$.
\end{enumerate}
The \term{$C'$-connected components} of $C$ are the equivalence classes of the $d$-cubes of $C$ under the
$C'$-connectedness relation.
\end{definition}

The recursive algorithm for constructing all maximum classes of VC-dimension
$d$ in the $n$-cube, detailed as Algorithm~\ref{alg:lifting}, considers each
possible $d$-maximum class $C$ in the $(n-1)$-cube and each possible
$(d-1)$-maximum subclass $C'$ of $C$ as the projection and reduction of a
$d$-maximum class in the $n$-cube, respectively.  The algorithm \term{lifts}
$C$ and $C'$ to all possible maximum classes in the $n$-cube. 
Then $C'\times\{0,1\}$ is contained in each lifted class; so all
that remains is to find the tails from the complement of the reduction in the
projection. It turns out that each $C'$-connected component $C_i$ of $C$ can
be lifted to either $C_i\times\{0\}$ or $C_i\times\{1\}$ arbitrarily and
independently of how the other $C'$-connected components are lifted. The
set of lifts equates to the set of $d$-maximum classes in the
$n$-cube that project-reduce to $(C,C')$.

\begin{algorithm}[tbp]
\caption{$\Construct{n}{d}$}
\begin{algorithmic}
\STATE \textbf{Given:} $n\in\N, d\in[n]$ \\
\STATE \textbf{Returns:} the set of $d$-maximum classes in $\{0,1\}^n$ \\[.4em]
\STATE {\small 1.}\hspace{0.2em} \textbf{if} $d=0$ \textbf{then} \textbf{return} $\left\{\left\{\vec{v}\right\}\mid\vec{v}\in\{0,1\}^n\right\}$\enspace;\\
\STATE {\small 2.}\hspace{0.2em} \textbf{if} $d=n$ \textbf{then} \textbf{return} $\{0,1\}^n$\enspace;\\
\STATE {\small 3.}\hspace{0.2em} $\mathcal{M} \leftarrow \emptyset$\enspace;\\
\STATE \hspace{1.0em} \textbf{for each} $C\in\Construct{n-1}{d}$, \\ \hspace{5.32em} $C'\in\Construct{n-1}{d-1}$ s.t. $C'\subset C$ \textbf{do} \\
\STATE {\small 4.}\hspace{1.0em} $\{C_1,\ldots,C_k\} \leftarrow $ $C'$-connected components of $C$\,;\\
\STATE {\small 5.}\hspace{1.0em} $\mathcal{M} \leftarrow \mathcal{M}\ \cup \bigcup_{\vec{p}\in\{0,1\}^k}\left\{\left(C'\times\{0,1\}\right)\cup\bigcup_{q\in[k]}C_q\times\{p_q\}\right\}$\,;\\
\STATE \hspace{1.0em} \textbf{done} \\
\STATE {\small 6.}\hspace{0.2em} \textbf{return} $\mathcal{M}$ ;\\
\end{algorithmic}
\label{alg:lifting}
\end{algorithm}

\begin{lemma}\label{lemma:lifting}
$\Construct{n}{d}$ (cf.~Algorithm~\ref{alg:lifting}) returns the set of maximum
classes of VC-dimension $d$ in the $n$-cube for all $n\in\N, d\in[n]$.
\end{lemma}

\begin{proof}
We proceed by induction on $n$ and $d$.  The base cases correspond to
$n\in\N, d\in\{0,n\}$ for which all maximum classes, enumerated
as singletons in the $n$-cube and the $n$-cube respectively, are
correctly produced by the algorithm. For the inductive step we assume that
for $n\in\N, d\in[n-1]$ all maximum classes of VC-dimension $d$ and $d-1$ in
the $(n-1)$-cube are already known by recursive calls to the algorithm.
Given this, we will show that $\Construct{n}{d}$ returns only $d$-maximum
classes in the $n$-cube, and that all such classes are produced by the
algorithm.

Let classes $C\in\Construct{n-1}{d}$ and $C'\in\Construct{n-1}{d-1}$ be such that
$C'\subset C$. Then $C$ is the union of a $d$-complete collection and
$C'$ is the union of a $(d-1)$-complete collection of cubes that are
faces of the cubes of $C$. 
Consider a concept class $C^\star$ formed from $C$
and $C'$ by Algorithm~\ref{alg:lifting}.  The algorithm partitions $C$ into
$C'$-connected components $C_1,\ldots,C_k$ each of which is a union of
$d$-cubes.  While $C'$ is lifted to
$C'\times\{0,1\}$, some subset of the components $\{C_i\}_{i\in S_0}$
are lifted to $\left\{C_i\times\{0\}\right\}_{i\in S_0}$ while the remaining
components 
are lifted to $\left\{C_i\times\{1\}\right\}_{i\notin S_0}$.  By definition
$C^\star$ is a $d$-complete collection of cubes with cardinality equal to 
$\smallchoose{n}{\leq d}$ since $|C^\star|=|C'|+|C|$~\citep{KW07}.  So
$C^\star$ is $d$-maximum~\citep[Theorem~34]{RBR08}.

If we now consider any $d$-maximum class $C^\star\subseteq\{0,1\}^n$, its
projection on $[n]\backslash\{i\}$ is a $d$-maximum class
$C\subseteq\{0,1\}^{n-1}$ and ${C^*}^i$ is the
$(d-1)$-maximum projection $C'\subset C$ of all the $d$-cubes in $C^\star$
which contain color $i$.  It is thus clear that $C^\star$ must be
obtained by lifting parts of the $C'$-connected components
of $C$ to the $1$ level and the remainder to the $0$ level, and $C'$ to 
$C'\times\{0,1\}$.  We will now show that if the vertices of each
component are not lifted to the same levels, then while the
number of vertices in the lift match that of a $d$-maximum class in the
$n$-cube, the number of edges are too few for such a maximum class.
Define a lifting operator on $C$ as $\ell(v) = \{v\}\times \ell_v$, where
$\ell_v \subseteq \{0,1\}$ and
\begin{eqnarray*}
|\ell_v| &=& \begin{cases}
2\enspace, & \mbox{if $v\in C'$} \\
1, & \mbox{if $v\in C\backslash C'$}
\end{cases}\enspace.
\end{eqnarray*}
Consider now an edge $\{u,v\}$ in $\Oinc{C}$.  By the definition of a
$C'$-connected component there exists some $C_j$ such that either
$u,v\in C_j\backslash C'$,  $u,v\in C'$ or WLOG
$u\in C_j\backslash C', v\in C'$.  In the first case
$\ell(u)\cup\ell(v)$ is an edge in the lifted graph iff $\ell_u=\ell_v$. In
the second case $\ell(u)\cup\ell(v)$ contains four edges and in the last it
contains a single edge.  Furthermore, it is clear that this accounts for all
edges in the lifted graph by considering the projection of an edge in the
lifted product.  Thus any lift other than those produced by
Algorithm~\ref{alg:lifting} induces strictly too few edges for a 
$d$-maximum class in the $n$-cube (cf.~\citealp[Corollary~7.5]{KW07}).
\end{proof}

\subsection{Corner-Peeling}

\citet[Conjecture~2]{KW07} conjectured that their
simple \term{Min-Peeling} procedure is a valid unlabeled compression
scheme for maximum classes.  Beginning with a concept 
class $C_0=C\subseteq\{0,1\}^n$, Min-Peeling operates by iteratively removing
a vertex $v_t$ of minimum-degree in $\Oinc{C_t}$ to produce
the peeled class $C_{t+1}=C_t\backslash\{v_t\}$.  The concept class
corresponding to $v_t$ is then represented by the dimensions  of the edges incident to
$v_t$ in $\Oinc{C_t}$, $\incidentdims{\Oinc{C_t}}{v_t}\subseteq[n]$. 
Providing that no-clashing holds for the algorithm, the
size of the min-peeling scheme is the largest degree encountered
during peeling.  Kuzmin and Warmuth predicted that this size is always at
most $d$ for $d$-maximum classes.  We explore these questions for a
related special case of peeling, where we prescribe which vertex 
to peel at step $t$ as follows.

\begin{definition}\label{def:peeling}
We say that $C\subseteq\{0,1\}^n$ can be \term{corner-peeled} if there exists
an ordering $v_1,\ldots,v_{|C|}$ of the vertices of $C$ such that, for each
$t\in[|C|]$ where $C_0=C$,
\begin{enumerate}
\item $v_t\in C_{t-1}$ and $C_t=C_{t-1}\backslash\{v_t\}$;
\item There exists a unique cube $C'_{t-1}$ of maximum dimension over all
cubes in $C_{t-1}$ containing $v_t$; 
\item The neighbors $\Gamma(v_t)$ of $v_t$ in $\Oinc{C_{t-1}}$ satisfy
$\Gamma(v_t)\subseteq C'_{t-1}$; and
\item $C_{|C|}=\emptyset$.
\end{enumerate}
The $v_t$ are termed the \term{corner vertices} of $C_{t-1}$ respectively.  If
$d$ is the maximum degree of each $v_t$ in $\Oinc{C_{t-1}}$, then $C$ is 
\term{$d$-corner-peeled}.
\end{definition}

Note that we do not constrain the cubes $C_t'$ to be of non-increasing
dimension. It turns out that an important property of maximum classes is
invariant to this kind of peeling.

\begin{definition}\label{def:geod-property}
We call a class $C\subseteq\{0,1\}^n$ \term{shortest-path closed} if 
for any $u,v\in C$, $\Oinc{C}$ contains a path connecting $u,v$ of length
$\|u - v\|_1$.
\end{definition}

\begin{lemma}\label{lemma:hamming-invariance}
If $C\subseteq\{0,1\}^n$ is shortest-path closed and $v\in C$ is a corner
vertex of $C$, then $C\backslash\{v\}$ is shortest-path closed.
\end{lemma}

\begin{proof}
Consider a shortest-path closed $C\subseteq\{0,1\}^n$.  Let $c$ be a corner
vertex of $C$, and denote the cube of maximum dimension in $C$, containing $c$, 
by $C'$.  Consider $\{u,v\}\subseteq C\backslash\{c\}$.  By assumption there
exists a $u$-$v$-path $p$ of length $\|u - v\|_1$ contained in $C$. If
$c$ is not in $p$ then $p$ is contained in the peeled product
$C\backslash\{c\}$.  If $c$ is in $p$ then $p$ must cross $C'$ such that there
is another path of the same length which avoids $c$, and thus
$C\backslash\{c\}$ is shortest-path closed.
\end{proof}

\subsubsection{Corner-Peeling Implies Compression}

\begin{theorem}\label{thm:corner-peeling-compresses}
If a maximum class $C$ can be corner-peeled then $C$ can be $d$-unlabeled
compressed.
\end{theorem}

\begin{proof}
The invariance of the shortest-path closed property under corner-peeling is
key. The corner-peeling unlabeled compression scheme represents each 
$v_t\in C$ by $r(v_t)=\incidentdims{\Oinc{C_{t-1}}}{v_t}$, the colors of the
cube $C'_{t-1}$ which is deleted from $C_{t-1}$ when $v_t$ is corner-peeled.
We claim that any two vertices $v_s,v_t\in C$ have non-clashing
representatives.  WLOG, suppose that $s<t$. 
The class $C_{s-1}$ must contain a shortest $v_s$-$v_t$-path $p$. Let
$i$ be the color of the single edge contained in $p$ that is incident to
$v_s$. Color $i$ appears once in $p$, and is contained in
$r(v_s)$.  This implies that $v_{s,i}\neq v_{t,i}$ and that
$i\in r(v_s)\cup r(v_t)$, and so
$v_s|\left(r(v_s)\cup r(v_t)\right)\neq v_t|\left(r(v_s)\cup r(v_t)\right)$.
By construction, $r(\cdot)$ is a bijection between $C$ and all subsets
of $[n]$ of cardinality $\le \VC{C}$.
\end{proof}

If the oriented one-inclusion graph, with each edge directed away from the
incident vertex represented by the edge's color, has no cycles, then
that representation's compression scheme is termed
\term{acyclic}~\citep{F89,BDL98,KW07}.

\begin{proposition}\label{prop:acyclic}
All corner-peeling unlabeled compression schemes are acyclic.
\end{proposition}

\begin{proof}
We follow the proof that the Min-Peeling Algorithm is
acyclic~\citep{KW07}.  Let $v_1,\ldots,v_{|C|}$ be a corner vertex ordering 
of $C$.  As a corner vertex $v_t$ is peeled, its unoriented incident edges
are oriented away from $v_t$.  Thus all edges incident to
$v_1$ are oriented away from $v_1$ and so the vertex cannot take part in any cycle.  For $t>1$
assume $V_t=\{v_s\mid s<t\}$ is disjoint from all cycles.  Then $v_t$ cannot be
contained in a cycle, as all incoming edges into $v_t$ are incident
to some vertex in $V_t$.  Thus the oriented $\Oinc{C}$ is indeed acyclic.
\end{proof}

\subsection{Boundaries of Maximum Classes}

We now turn to the geometric boundaries of maximum classes, which are closely
related to corner-peeling.

\begin{definition}\label{def:boundaries}
The boundary $\partial C$ of a $d$-maximum class $C$ is defined as all the
$(d-1)$-subcubes which are the faces of a single $d$-cube in $C$.
\end{definition}

Maximum classes, when viewed as cubical complexes, are analogous to
soap films (an example of a minimal energy surface encountered in nature), which
are obtained when a wire frame is dipped into a soap solution. Under this
analogy the boundary corresponds to the wire frame and the number of $d$-cubes
can be considered the area of the soap film. An important property of the
boundary of a maximum class is that all lifted reductions meet the boundary  
multiple times.

\begin{theorem}\label{thm:boundary-cubes}
Every $d$-maximum class has boundary containing at
least two $(d-1)$-cubes of every combination of $d-1$ colors, for all $d>1$. 
\end{theorem}

\begin{proof}
We use the lifting construction of Section~\ref{sec:lifting}. Let
$C^\star\subseteq\{0,1\}^n$ be a $2$-maximum class and consider color
$i\in[n]$. Then the reduction ${C^\star}^i$ is an unrooted
tree with at least two leaves, each of which lifts to an $i$-colored edge in
$C^\star$.  Since the leaves are of degree $1$ in ${C^\star}^i$, the
corresponding lifted edges belong to exactly one $2$-cube in $C^\star$ and so
lie in $\partial C^\star$. Consider now a $d$-maximum class
$C^\star\subseteq\{0,1\}^n$ for $d>2$, and make the inductive assumption that
the projection $C=\proj{[n-1]}{C^\star}$ has two of each type of $(d-1)$-cube,
and that the reduction $C'={C^\star}^{n}$ has two of each type of
$(d-2)$-cube, in their boundaries. Pick $d-1$ colors $I\subseteq[n]$.  If
$n\in I$ then consider two $(d-2)$-cubes colored by $I\backslash\{x_n\}$ in
$\partial C'$.  By the same argument as in the base case, these lift to two
$I$-colored cubes in $\partial C^\star$. If $n\notin I$ then
$\partial C$ contains two $I$-colored $(d-1)$-cubes.  For each cube, if the
cube is contained in $C'$ then it has two lifts one of which
is contained in $\partial C^\star$, otherwise its unique lift is contained
in $\partial C^\star$.  Therefore $\partial C^\star$ contains at least two
$I$-colored cubes.
\end{proof}

Having a large boundary is an important property of maximum classes that does
not follow from contractibility.

\begin{figure}[t]
\begin{center}
\begin{minipage}[t]{1\textwidth}
\centering
\includegraphics[width=0.7\textwidth]{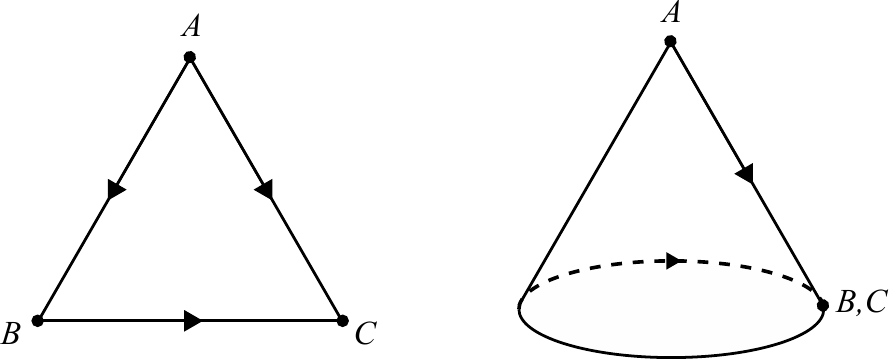}
\caption{The first steps of building the dunce hat in Example~\ref{eg:dunce}.}
\label{fig:dunce}
\end{minipage}
\end{center}
\end{figure}

\begin{example}\label{eg:dunce}
Take a $2$-simplex with vertices $A,B,C$. Glue the edges $AB$ to $AC$ to form a cone. Next glue the end loop $BC$ to the edge $AB$ . The result is a complex $D$ with a single vertex, edge and $2$-simplex, which is classically known as the
\term{dunce hat} (cf.~Figure~\ref{fig:dunce}). It is not hard to verify that
$D$ is contractible, but has no (geometric) boundary.
\end{example}

Although Theorem~\ref{thm:boundary-cubes} 
will not be explicitly used in the sequel, we return to 
boundaries of maximum complexes later.

\section{Euclidean Arrangements}
\label{sec:euclidean}

\begin{definition}
A \term{linear arrangement} is a collection of $n\geq d$ oriented hyperplanes in
$\R^d$. Each region or cell in the complement of the arrangement 
is naturally associated with a concept in $\{0,1\}^n$;
the side of the $\nth{i}{th}$ hyperplane on which a cell falls determines
the concept's $\nth{i}{th}$ component.
A \term{simple arrangement} is a linear arrangement in which any subset of
$d$ planes has a 
unique point in common and all subsets of $d+1$ planes have an empty mutual
intersection.  Moreover any subset of $k<d$ planes meet in a plane of dimension $d-k$.
Such a collection of $n$ planes is also said to be in 
\term{general position}.
\end{definition}

Many of the familiar operations on concept classes in the $n$-cube have
elegant analogues on arrangements.
\begin{itemize}
\item Projection on $[n]\backslash\{i\}$ corresponds to removing the
$\nth{i}{th}$ plane;
\item The reduction $C^i$ is
the new arrangement given by the intersection of $C$'s arrangement with the $\nth{i}{th}$ plane; 
and
\item The corresponding lifted reduction is the collection of cells in the 
arrangement that adjoin the $\nth{i}{th}$ plane.
\end{itemize}
A $k$-cube in the one-inclusion graph corresponds to a collection of $2^k$ cells, all having
a common $(d-k)$-face, which is contained in the intersection of $k$ planes, and an edge corresponds to a pair of cells which have a common face on a single plane.  The following result is due
to~\citet{E87}.

\begin{lemma}\label{lemma:simple-is-maximum}
The concept class $C\subseteq\{0,1\}^n$ induced by a simple linear arrangement
of $n$ planes in $\R^d$ is $d$-maximum.
\end{lemma}

\begin{proof}
Note that $C$ has VC-dimension at most $d$, since general position is
invariant to projection i.e., no $d+1$ planes are shattered.  Since
$C$ is the union of a $d$-complete collection of cubes (every cell contains
$d$-intersection points in its boundary) it follows that $C$ is
$d$-maximum~\citep{RBR08}.
\end{proof}

\begin{corollary}\label{cor:generic-intersect-is-submax}
Let $A$ be a simple linear arrangement of $n$ hyperplanes in $\R^d$ with
corresponding $d$-maximum $C\subseteq\{0,1\}^n$.  The intersection of $A$ with
a generic hyperplane corresponds to a $(d-1)$-maximum class
$C'\subseteq C$.  In particular if all $d$-intersection points of $A$ lie to
one side of the generic hyperplane, then $C'$ lies on the boundary of $C$; and
$\partial C$ is the disjoint union of two $(d-1)$-maximum sub-classes.
\end{corollary}

\begin{proof}
The intersection of $A$ with a generic hyperplane is again a simple arrangement
of $n$ hyperplanes but now in $\R^{d-1}$.  Hence by
Lemma~\ref{lemma:simple-is-maximum} $C'$ is a $(d-1)$-maximum class in the
$n$-cube.  $C'\subseteq C$ since the adjacency relationships on the cells of 
the intersection are inherited from those of $A$.

Suppose that all $d$-intersections in $A$ lie in one half-space of the generic
hyperplane.  $C'$ is the union of a $(d-1)$-complete collection.  We claim that
each of these $(d-1)$-cubes is a face of exactly one $d$-cube in $C$ and is 
thus in $\partial C$.  A $(d-1)$-cube in $C'$ corresponds to a line in $A$
where $d-1$ planes mutually intersect.  The $(d-1)$-cube is a face of a
$d$-cube in $C$ iff this line is further intersected by a $\nth{d}{th}$ plane.
This occurs for exactly one plane, which is closest to the generic hyperplane
along this intersection line.
For once the $d$-intersection point is reached, when following along the line
away from the generic plane, a new cell is entered.  This verifies the second
part of the result.

Consider two parallel generic hyperplanes $h_1,h_2$ such that all
$d$-intersection points of $A$ lie in between them.  We claim that each 
$(d-1)$-cube in $\partial C$ is in exactly one of the concept classes induced
by the intersection of $A$ with $h_1$ and $A$ with $h_2$. Consider an
arbitrary $(d-1)$-cube in $\partial C$.
As before this cube corresponds to a region of a line formed by a mutual
intersection of $d-1$ planes.  Moreover this region is a ray, with one end-point
at a $d$-intersection.  Because the ray begins at a point between the
generic hyperplanes $h_1,h_2$, it follows that the ray must cross exactly one
of these.
\end{proof}

\begin{corollary}\label{cor:simple-is-ball}
Let $A$ be a simple linear arrangement of $n$ hyperplanes in $\R^d$ and let
$C\subseteq\{0,1\}^n$ be the corresponding $d$-maximum class.  Then $C$
considered as a cubical complex is homeomorphic to the $d$-ball $B^d$; and 
$\partial C$ considered as a $(d-1)$-cubical complex is homeomorphic to the
$(d-1)$-sphere $S^{d-1}$.
\end{corollary}

\begin{proof}
We construct a Voronoi cell decomposition corresponding to the set of $d$-intersection points inside a very large ball in Euclidean space. By induction on $d$, we claim that this is a cubical complex and the vertices and edges correspond to the class $C$. By induction, on each hyperplane, the induced arrangement has a Voronoi cell decomposition which is a $(d-1)$-cubical complex with edges and vertices matching the one-inclusion graph for the tail of $C$ corresponding to the label associated with the hyperplane. It is not hard to see that the Voronoi cell defined by a $d$-intersection point $p$ on this hyperplane is a $d$-cube. In fact, its $(d-1)$-faces correspond to the Voronoi cells for $p$, on each of the $d$ hyperplanes passing through $p$. We also see that this $d$-cube has a single vertex in the interior of each of the $2^d$ cells of the arrangement adjacent to $p$. In this way, it follows that the vertices of this Voronoi cell decomposition are in bijective correspondence to the cells of the hyperplane arrangement. Finally the edges of the Voronoi cells pass through the faces in the hyperplanes. So these correspond bijectively to the edges of $C$, as there is one edge for each face of the hyperplanes. Using a very large ball, containing all the $d$-intersection points, the boundary faces become spherical cells. In fact, these form a spherical Voronoi cell decomposition, so it is easy to replace these by linear ones by taking the convex hull of their vertices. So a piecewise linear cubical complex $\bf C$ is constructed, which has one-skeleton (graph consisting of all vertices and edges) isomorphic to the one-inclusion graph for $C$. 

 Finally we want to prove that $\bf C$ is homeomorphic to $B^d$. This is quite easy by construction. For we see that $\bf C$ is obtained by dividing up $B^d$ into Voronoi cells and replacing the spherical boundary cells by linear ones, using convex hulls of the boundary vertices. This process is clearly given by a homeomorphism by projection. In fact, the homeomorphism preserves the PL-structure so is a PL homeomorphism. 
\end{proof}

The following example demonstrates that not all maximum classes of VC-dimension
$d$ are homeomorphic to the $d$-ball.  The key to such examples is branching.

\begin{figure}[t]
\begin{center}
\begin{tabular}{l|cccc}
& $x_1$ & $x_2$ & $x_3$ & $x_4$ \\
\hline
$v_0$ & 0 & 0 & 0 & 0 \\
$v_1$ & 1 & 0 & 0 & 0 \\
$v_2$ & 0 & 1 & 0 & 0 \\
$v_3$ & 0 & 0 & 1 & 0 \\
$v_4$ & 1 & 0 & 1 & 0 \\
$v_5$ & 1 & 1 & 0 & 0 \\
$v_6$ & 0 & 1 & 1 & 0 \\
$v_7$ & 1 & 0 & 0 & 1 \\
$v_8$ & 1 & 1 & 0 & 1 \\
$v_9$ & 0 & 1 & 0 & 1 \\
$v_{10}$ & 0 & 1 & 1 & 1 \\
\end{tabular}
\end{center}
\caption{A $2$-maximum class in $\{0,1\}^4$ having a simple linear line
arrangement in $\R^2$.}
\label{tab:euclidean}
\end{figure}

\begin{example}
A simple linear arrangement in $\R$ corresponds to points on the line---cells
are simply intervals between these points and so corresponding $1$-maximum
classes are sticks.  Any tree that is not a stick can therefore not be represented
as a simple linear arrangement in $\R$ and is also not homeomorphic to the
$1$-ball which is simply the interval $[-1,1]$.
\end{example}

As \citet{KW07} did previously, consider a generic
hyperplane $h$ sweeping across a simple linear arrangement $A$.  
$h$ begins with all $d$-intersection points of $A$ lying in its positive half-space $\H_+$.  
The concept corresponding to
cell $c$ is peeled from $C$ when $|\H_+\cap c|=1$, i.e., $h$ 
crosses the last $d$-intersection point adjoining $c$.  At any step in
the process, the result of peeling $j$ vertices from $C$ to reach $C_j$, is
captured by the arrangement $\H_+\cap A$ for the appropriate $h$.  

\begin{figure}[t]
\begin{center}
\begin{minipage}[t]{.47\textwidth}
\centering
\includegraphics[width=0.9\textwidth]{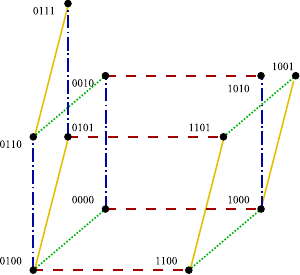}
\caption{The $2$-maximum class in the $4$-cube, enumerated in
Figure~\ref{tab:euclidean}.}
\label{fig:euclidean-ncube}
\end{minipage}\hspace{1.3em}
\begin{minipage}[t]{.47\textwidth}
\centering
\includegraphics[width=0.9\textwidth]{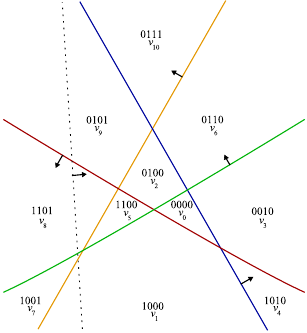}
\caption{A simple linear line arrangement corresponding to the class in
Figure~\ref{tab:euclidean}, swept by the dashed line.  Each cell has a unique vertex.}
\label{fig:euclidean-arrange}
\end{minipage}
\end{center}
\end{figure}

\begin{example}\label{eg:euclidean}
Figure~\ref{tab:euclidean} enumerates the 11 vertices of a $2$-maximum class in
the $4$-cube.  Figures~\ref{fig:euclidean-arrange} and~\ref{fig:euclidean-ncube}
display a hyperplane arrangement in Euclidean space and its Voronoi cell decomposition, 
corresponding to this maximum class.  In this case, sweeping the vertical
dashed line across the arrangement corresponds to a partial
corner-peeling of the concept class with peeling sequence $v_7$, $v_8$, $v_5$,
$v_9$, $v_2$, $v_0$.  What remains is the $1$-maximum stick
$\{v_1, v_3, v_4, v_6, v_{10}\}$. 
\end{example}

Next we resolve the first half of~\citep[Conjecture~1]{KW07}.

\begin{theorem}\label{thm:sweeping-linear-is-corner}
Any $d$-maximum class $C\subseteq\{0,1\}^n$ corresponding to a simple linear
arrangement $A$ can be corner-peeled by sweeping $A$, and this process is a 
valid unlabeled compression scheme for $C$ of size $d$.
\end{theorem}

\begin{proof}
We must show that as the $\nth{j}{th}$ $d$-intersection point $p_j$ is crossed, there is a
corner vertex of $C_{j-1}$ peeled away.  It then follows that sweeping a generic hyperplane $h$
across $A$ corresponds to corner-peeling $C$ to a $(d-1)$-maximum sub-class
$C'\subseteq\partial C$ by Corollary~\ref{cor:generic-intersect-is-submax}. 
Moreover $C'$ corresponds to a simple linear arrangement of $n$ hyperplanes
in $\R^{d-1}$.  

We proceed by induction on $d$, noting that for $d=1$ corner-peeling is trivial.
Consider $h$ as it approaches the $\nth{j}{th}$ $d$-intersection point $p_j$. 
The $d$ planes defining this point intersect $h$ in a simple arrangement of
hyperplanes on $h$.  There is a compact cell $\Delta$ for the arrangement on $h$, which is a
$d$-simplex\footnote{A topological simplex---the convex hull of
$d+1$ affinely independent points in $\R^{d}$.} and shrinks to
a point as $h$ passes through $p_j$. 
We claim that the cell $c$ for the arrangement $A$, whose intersection with $h$ is $\Delta$,
is a corner vertex $v_j$ of $C_{j-1}$. Consider the lines formed by intersections
of $d-1$ of the $d$ hyperplanes, passing through $p_j$. Each is a segment starting at $p_j$ and
ending at $h$ without passing through any other $d$-intersection points.  So all faces of 
hyperplanes adjacent to $c$ meet $h$ in faces of $\Delta$.
Thus, there are no edges in $C_{j-1}$ starting at the vertex corresponding to $p_j$,
except for those in the cube $C_{j-1}'$, which consists of all cells adjacent to $p_j$ in the arrangement $A$. 
So $c$ corresponds to a corner vertex $v_j$ of the $d$-cube $C_{j-1}'$ in $C_{j-1}$. Finally, just after the simplex is a point, $c$ is no longer in $\H_+$ and so
$v_j$ is corner-peeled from $C_{j-1}$.

Theorem~\ref{thm:corner-peeling-compresses} completes the proof that this
corner-peeling of $C$ constitutes unlabeled compression.
\end{proof}

\begin{corollary}\label{cor:euclidean-nondec}
The sequence of cubes $C'_0,\ldots,C'_{|C|}$, removed when corner-peeling 
by sweeping simple linear arrangements, is of non-increasing dimension. In fact, there are 
$n \choose d$ cubes of dimension $d$, then $n \choose d-1$ cubes of dimension $d-1$, etc. 
\end{corollary}

While corner-peeling and min-peeling share some properties in common, they are
distinct procedures.

\begin{example}\label{eg:cpeel-mpeel-different}
Consider sweeping a simple linear arrangement corresponding to a $2$-maximum class.
After all but one $2$-intersection point has been swept, the corresponding
corner-peeled class $C_t$ is the union of a single $2$-cube with a $1$-maximum
stick.  Min-peeling applied to $C_t$ would first peel a leaf, while 
corner-peeling must begin with the $2$-cube.
\end{example}

The next result follows from our counter-examples to Kuzmin \& Warmuth's
minimum degree conjecture~\citep{RBR08}.

\begin{corollary}
There is no constant $c$ so that all maximal classes of VC dimension $d$
can be embedded into maximum classes corresponding to simple hyperplane
arrangements of dimension $d+c$.
\end{corollary}

\section{Hyperbolic Arrangements}

We briefly discuss the Klein model of hyperbolic geometry~\citep[pg. 7]{R94}.
Consider the open unit ball $\bbH^k$ in $\R^k$. Geodesics (lines of
shortest length in the geometry) are given by intersections of straight lines
in $\R^k$ with the unit ball. Similarly planes of any dimension between $2$
and $k-1$ are given by intersections of such planes in $\R^k$ with the unit
ball. Note that such planes are completely determined by their spheres of
intersection with the unit sphere $S^{k-1}$, which is called the ideal
boundary of hyperbolic space $\bbH^k$. Note that in the appropriate metric,
the ideal boundary consists of points which are infinitely far from all
points in the interior of the unit ball. 

We can now see immediately that a simple hyperplane arrangement in $\bbH^k$
can be described by taking a simple hyperplane arrangement in $\R^k$ and
intersecting it with the unit ball. However we require an important additional
property to mimic the Euclidean case. Namely we add the constraint that
every subcollection of $d$ of the hyperplanes in $\bbH^k$ has
mutual intersection points inside $\bbH^k$, and that no $(d+1)$-intersection
point lies in $\bbH^k$. We need this requirement to obtain that the
resulting class is maximum.

\begin{definition}\label{def:simple-hyperbolic}
A \term{simple hyperbolic $d$-arrangement} is a collection of $n$ hyperplanes in
$\bbH^k$ with the property that every sub-collection of $d$ hyperplanes
mutually intersect in a $(k-d)$-dimensional hyperbolic plane, and that every
sub-collection of $d+1$ hyperplanes mutually intersect as the empty set.
\end{definition}

\begin{corollary}\label{cor:simple-hyperbolic-is-maximum}
The concept class $C$ corresponding to a simple $d$-arrangement of hyperbolic
hyperplanes in $\bbH^k$ is $d$-maximum in the $k$-cube.
\end{corollary}

\begin{proof}
The result follows by the same argument as before.  Projection cannot
shatter any $(d+1)$-cube and the class is a
complete union of $d$-cubes, so is $d$-maximum.
\end{proof}

The key to why hyperbolic arrangements represent many new
maximum classes is that they allow flexibility of choosing $d$ and $k$
independently. This is significant because the unit ball can be chosen to
miss much of the intersections of the hyperplanes in Euclidean space. Note
that the new maximum classes are embedded in maximum
classes induced by arrangements of linear hyperplanes in Euclidean space.

A simple example is any $1$-maximum class. It is easy to see that
this can be realized in the hyperbolic plane by choosing an
appropriate family of lines and the unit ball in the appropriate position.
In fact, we can choose sets of pairs of points on the unit circle, which
will be the intersections with our lines. So long as these pairs of points
have the property that the smaller arcs of the circle between them are
disjoint, the lines will not cross inside the disk and the desired
$1$-maximum class will be represented. 

Corner-peeling maximum classes represented by hyperbolic hyperplane
arrangements proceeds by sweeping, just as in the Euclidean case. Note first
that intersections of the hyperplanes of the arrangement with
the moving hyperplane appear precisely when there is a first intersection at the ideal
boundary. Thus it is necessary to slightly perturb the collection of hyperplanes to ensure that
only one new intersection with the moving hyperplane occurs at any time.  
Note also that new intersections of the sweeping hyperplane 
with the various lower dimensional planes of intersection 
between the hyperplanes appear similarly at the ideal boundary. The important
claim to check is that the intersection at the ideal boundary between the
moving hyperplane and a lower dimensional plane, consisting entirely of $d$ intersection points,
corresponds to a corner-peeling move. We include two examples to illustrate
the validity of this claim.

\begin{example}
In the case of a $1$-maximum class coming from disjoint lines in $\bbH^2$, 
a cell can disappear when the sweeping hyperplane meets a line at an ideal
point. This cell is indeed a vertex of the tree, i.e., a corner-vertex.
\end{example}

\begin{example}
Assume that we have a family of $2$-planes in the unit $3$-ball which meet in pairs
in single lines, but there are no triple points of intersection, corresponding to
a $2$-maximum class. A corner-peeling move occurs when a region bounded by two half disks 
and an interval disappears, in the positive half space bounded by the sweeping hyperplane. 
Such a region can be visualized by taking a slice out of an orange. Note that the final point of contact between the hyperplane and the
 region is at the end of a line of intersection between two planes on the ideal boundary. 
\end{example}

\begin{figure}[t]
\begin{center}
\begin{minipage}[t]{.47\textwidth}
\centering
\includegraphics[width=0.9\textwidth]{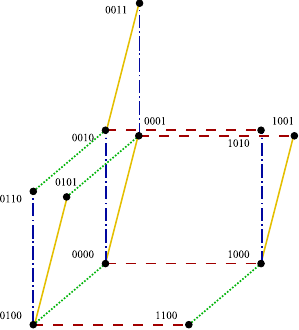}\\[0.5em](a)
\end{minipage}\hspace{1.3em}
\begin{minipage}[t]{.47\textwidth}
\centering
\includegraphics[width=0.9\textwidth]{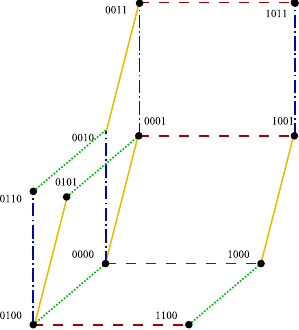}\\[0.5em](b)
\end{minipage}
\caption{$2$-maximum classes in $\{0,1\}^4$ that can be represented as 
hyperbolic arrangements but not as Euclidean arrangements.}
\label{fig:hyperbolic-ncubes}
\end{center}
\end{figure}

We next observe that sweeping by generic hyperbolic hyperplanes induces 
corner-peeling of the corresponding maximum class, extending
Theorem~\ref{thm:sweeping-linear-is-corner}. As the generic hyperplane
sweeps across hyperbolic space, not only do 
swept cells correspond to corners of $d$-cubes but also
to corners of lower dimensional cubes as well.  Moreover, the order of
the dimensions of the cubes which are
corner-peeled can be arbitrary---lower dimensional cubes may be
corner-peeled before all the higher dimensional cubes are corner-peeled.  This
is in contrast to Euclidean sweepouts (cf.~Corollary~\ref{cor:euclidean-nondec}).  Similar to Euclidean sweepouts,
hyperbolic sweepouts correspond to corner-peeling and not min-peeling.

\begin{theorem}\label{thm:sweeping-hyperbolic-is-corner}
Any $d$-maximum class $C\subseteq\{0,1\}^n$ corresponding to a simple hyperbolic
$d$-arrangement $A$ can be corner-peeled by sweeping $A$ with a generic hyperbolic
hyperplane.
\end{theorem}

\begin{proof}
We follow the same strategy of the proof of
Theorem~\ref{thm:sweeping-linear-is-corner}.  For sweeping in hyperbolic
space $\bbH^k$, the generic hyperplane $h$ is initialized as tangent to
$\bbH^k$.  As $h$ is swept across $\bbH^k$, new intersections appear with $A$
just after $h$ meets the non-empty intersection of a subset of hyperplanes of
$A$ with the ideal boundary.  Each $d$-cube $C'$ in $C$ still corresponds to
the cells adjacent to the intersection $I_{C'}$ of $d$ hyperplanes. But now $I_{C'}$ is a
($k-d$)-dimensional hyperbolic hyperplane.  A cell $c$ adjacent to $I_{C'}$
is corner-peeled  precisely when $h$ last intersects $c$ at a point of $I_{C'}$ at the ideal boundary.  
As for simple linear arrangements, the general position of $A\cup\{h\}$ ensures that
corner-peeling events never occur simultaneously.  For the case $k=d+1$, as for
the simple linear arrangements just prior to the corner-peeling of $c$,
$\H_+\cap c$ is homeomorphic to a $(d+1)$-simplex with a missing face on
the ideal boundary.  And so as in the simple linear case, this $d$-intersection
point corresponds to a corner $d$-cube.  In the case $k>d+1$, $\H_+\cap c$
becomes a $(d+1)$-simplex (as before) multiplied by $\R^{k-d-1}$. If $k=d$, then 
the main difference is just before corner-peeling of $c$, $\H_+\cap c$ is 
homeomorphic to a $k$-simplex  which may be either closed (hence in the interior of $\bbH^k$)
or with a missing face on the ideal boundary. The rest of the argument
remains the same, except for one important observation. 

Although swept corners in hyperbolic arrangements can be of cubes of differing
dimensions, these dimensions never exceed $d$ and so the proof that
sweeping simple linear arrangements induces $d$-compression schemes is still
valid.
\end{proof}

\begin{figure}[t]
\begin{center}
\begin{minipage}[t]{.47\textwidth}
\centering
\includegraphics[width=0.9\textwidth]{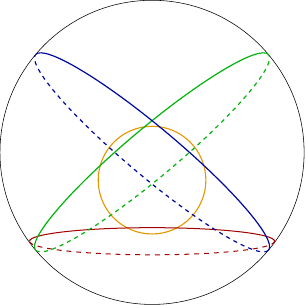}\\[0.5em](a)
\end{minipage}\hspace{1.3em}
\begin{minipage}[t]{.47\textwidth}
\centering
\includegraphics[width=0.9\textwidth]{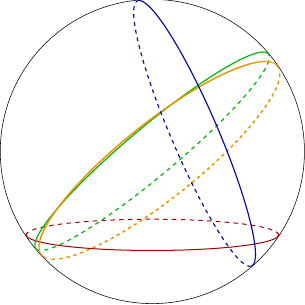}\\[0.5em](b)
\end{minipage}
\caption{Hyperbolic hyperplane arrangements corresponding to the
classes in Figure~\ref{fig:hyperbolic-ncubes}. In both cases the four
hyperbolic planes meet in 6 straight line segments (not shown). The 
planes' colors correspond to the edges' colors in Figure~\ref{fig:hyperbolic-ncubes}.}
\label{fig:hyperbolic-arranges}
\end{center}
\end{figure}

\begin{figure}[t]
\begin{center}
\begin{minipage}[t]{.47\textwidth}
\centering
\includegraphics[width=0.9\textwidth]{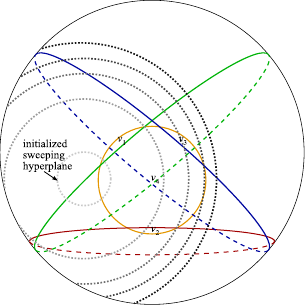}
\caption{The simple hyperbolic arrangement of
Figure~\ref{fig:hyperbolic-arranges}.(a) with a generic sweeping hyperplane
shown in several positions before and after it sweeps past four cells.}
\label{fig:hyperbolic1-sweep-arrange}
\end{minipage}\hspace{1.3em}
\begin{minipage}[t]{.47\textwidth}
\centering
\includegraphics[width=0.9\textwidth]{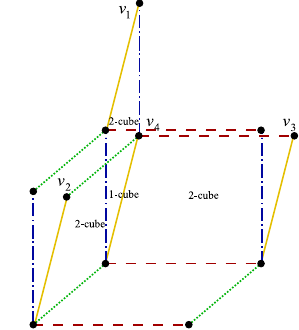}
\caption{The $2$-maximum class in $\{0,1\}^4$ of
Figure~\ref{fig:hyperbolic-ncubes}.(a), with the first four corner-vertices
peeled by the hyperbolic arrangement sweeping of
Figure~\ref{fig:hyperbolic1-sweep-arrange}.  Notice that three $2$-cubes are 
peeled, then a $1$-cube (all shown) followed by $2$-cubes.}
\label{fig:hyperbolic1-sweep-ncube}
\end{minipage}
\end{center}
\end{figure}

\begin{example}
Constructed with lifting, Figure~\ref{fig:hyperbolic-ncubes} completes the
enumeration, up to symmetry, of the $2$-maximum classes in $\{0,1\}^4$ begun
with Example~\ref{eg:euclidean}.  These cases cannot be
represented as simple Euclidean linear arrangements, since their boundaries do not satisfy the condition of  Corollary~\ref{cor:simple-is-ball} but can be represented as 
hyperbolic arrangements as in Figure~\ref{fig:hyperbolic-arranges}.
Figures~\ref{fig:hyperbolic1-sweep-arrange}
and~\ref{fig:hyperbolic1-sweep-ncube}
display the sweeping of a general hyperplane across the former arrangement and
the corresponding corner-peeling.  Notice that the corner-peeled cubes'
dimensions decrease and then increase.
\end{example}

\begin{corollary}
There is no constant $c$ so that all maximal classes of VC dimension $d$ can
be embedded into maximum classes corresponding to simple hyperbolic hyperplane
arrangements of VC dimension $d+c$.
\end{corollary}

This result follows from our counter-examples to Kuzmin \& Warmuth's
minimum degree conjecture~\citep{RBR08}.

Corollary~\ref{cor:simple-hyperbolic-is-maximum} gives a proper superset of
simple linear hyperplane arrangement-induced maximum classes as hyperbolic
arrangements. We will prove in Section~\ref{sec:PL-arrangements} that all
maximum classes
can be represented as PL hyperplane arrangements in a ball. These are the
topological analogue of hyperbolic hyperplane arrangements. If the boundary
of the ball is removed, then we obtain an arrangement of PL hyperplanes in
Euclidean space.

\section{Infinite Euclidean and Hyperbolic Arrangements}

We consider a simple example of an infinite maximum class which admits
corner-peeling and a compression scheme analogous to those of previous
sections.

\begin{example}\label{eg:infinite1}
Let $\mathcal L$ be the set of lines in the plane of the form
$L_{2m} = \{(x,y)\mid x=m\}$ and $L_{2n+1} = \{(x,y)\mid y=n\}$ for
$m,n\in\N$. Let $v_{00}$, $v_{0n}$, $v_{m0}$, and $v_{mn}$ be the cells bounded
by the lines $\{L_2, L_3\}$, $\{L_2, L_{2n+1}, L_{2n+3}\}$, 
$\{L_{2m}, L_{2m+2}, L_3\}$, and $\{L_{2m}, L_{2m+2},L_{2n+1}, L_{2n+3}\}$,
respectively. Then the cubical complex $C$, with
vertices $v_{mn}$, can be
corner-peeled and hence compressed, using a sweepout by the lines
$\{(x,y)\mid x+(1 + \epsilon)y = t\}$ for $t \ge 0$ and any small fixed
irrational $\epsilon >0$. $C$ is a $2$-maximum class and the
unlabeled compression scheme is also of size $2$.
\end{example}

To verify the properties of this example, notice that sweeping as specified
corresponds to corner-peeling the vertex
$v_{00}$, then the vertices $v_{10}, v_{01}$,  then the remaining vertices
$v_{mn}$. The lines $x+(1+\epsilon)y=t$ are
generic as they pass through only one intersection point of $\mathcal L$
at a time. Additionally, representing $v_{00}$ by
$\emptyset$, $v_{0n}$ by $\{L_{2n+1}\}$, $v_{m0}$ by $\{L_{2m}\}$ and 
$v_{mn}$ by $\{L_{2m}, L_{2n+1}\}$ constitutes a valid unlabeled
compression scheme.  
Note that the compression scheme is associated with sweeping across the
arrangement in the direction of decreasing $t$.
This is necessary to pick up the boundary vertices of $C$ last in the
sweepout process, so that they have either singleton representatives or the
empty set. In this way, similar to~\citet{KW07}, we obtain a compression scheme
so that every labeled sample of size $2$ is associated with a unique concept in
$C$, which is consistent with the sample. On the other hand to obtain
corner-peeling, we need the
sweepout to proceed with $t$ increasing so that we can begin at the boundary
vertices of $C$.

In concluding this brief discussion, we note that many infinite collections
of simple hyperbolic hyperplanes and Euclidean hyperplanes can also be
corner-peeled and compressed, even if intersection points and cells
accumulate. However a key requirement in the Euclidean case is that the
concept class $C$ has a non-empty boundary, when considered as a cubical
complex. An easy approach is to assume that all the
$d$-intersections of the arrangement lie in  a half-space.
Moreover, since the boundary must also admit corner-peeling, we require
more conditions, similar to having all the intersection points lying in
an octant. 

\begin{example}
In $\R^3$, choose the family of planes $\mathcal P$ of
the form $P_{3n+i} = \{\vec{x}\in\R^3\mid x_{i+1} = 1 - {1 / n}\}$ for
$ n \ge 1$ and $i\in\{0,1,2\}$. A corner-peeling scheme is induced by
sweeping a generic plane 
$\{\vec{x}\in\R^3\mid x_1 + \alpha x_2 + \beta x_3 = t\}$ across the
arrangement, where $t$ is a parameter and $1,\alpha, \beta$ are algebraically independent (in particular, no integral linear combination is rational)
and $\alpha, \beta$ are both close to $1$. This example has similar properties to Example~\ref{eg:infinite1}: the compression scheme is again given by decreasing $t$ whereas corner-peeling
corresponds to increasing $t$. Note that cells shrink to points, as
$\vec{x} \rightarrow \vec{1}$ and the volume of cells converge to
zero as $n \rightarrow \infty$, or equivalently any 
$x_i \rightarrow 1$. 
\end{example}

\begin{example}
In the hyperbolic plane $\bbH^2$, represented as the unit circle centered at the origin in $\R^2$,
choose the family of lines $\mathcal L$
given by $L_{2n} = \{(x,y)\mid x=1 - {1 / n}\}$ and
$L_{2n+1} = \{(x,y)\mid x+ny=1\} $, for $n \ge 1$. This arrangement has
corner-peeling and compression schemes given by sweeping across 
$\mathcal L$ using the generic line  $\{y =t\}$.
\end{example}

\section{Piecewise-Linear Arrangements}\label{sec:PL-arrangements}

A \term{PL hyperplane} is the image of a proper
piecewise-linear homeomorphism from the $(k-1)$-ball $B^{k-1}$ into $B^k$, i.e., the inverse image of the boundary $S^{k-1}$ of the $k$-ball is $S^{k-2}$,~\citep{RS82}.
A \term{simple PL $d$-arrangement} is an arrangement of $n$ PL hyperplanes
such that every subcollection of $j$ hyperplanes meet transversely in a
$(k-j)$-dimensional PL plane for $2\leq j\leq d$ and every subcollection of
$d+1$ hyperplanes are disjoint. 

\subsection{Maximum Classes are Represented by Simple PL Hyperplane
Arrangements}

Our aim is to prove the following theorem by a series of steps. 

\begin{theorem}\label{thm:maximum-is-PL} 
Every $d$-maximum class $C\subseteq\{0,1\}^n$ can be represented by a
simple arrangement
of PL hyperplanes in an $n$-ball. Moreover the corresponding simple
arrangement of PL hyperspheres in the $(n-1)$-sphere also represents $C$,
so long as $n>d+1$.
\end{theorem}

\subsubsection{Embedding a $d$-Maximum Cubical Complex in the $n$-cube into an $n$-ball.}

We begin with a $d$-maximum cubical complex $C\subseteq\{0,1\}^n$ embedded
into $[0,1]^n$.  This gives a natural embedding of $C$ into $\R^n$. Take a small
regular neighborhood $\cN$ of $C$ so that the boundary $\partial\cN$ of
$\cN$ will be a closed manifold of dimension $n-1$.  Note that $\cN$ is
contractible because it has a deformation retraction onto $C$ and so $\partial\cN$
is a homology $(n-1)$-sphere (by a standard, well-known argument from
topology due to~\citealt{M61}).  Our aim is to prove that $\partial\cN$ is an
$(n-1)$-sphere and $\cN$ is an $n$-ball.  There are two ways of proving this:
show that $\partial\cN$ is simply connected and invoke the well-known solution to
the generalised Poincar\'e conjecture~\citep{S61}, or  use the cubical
structure of the $n$-cube and $C$ to directly prove the result. We adopt the latter
approach, although the former works
fine. The advantage of the latter is that it produces the required
hyperplane arrangement, not just the structures of  $\partial\cN$ and 
$\cN$. 

\subsubsection{Bisecting Sets}

\begin{figure}[t]
\begin{center}
\centering
\includegraphics[width=0.4\textwidth]{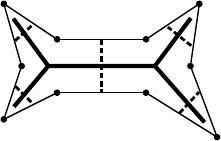}
\caption{A 1-maximum class (thick solid lines) with its fattening (thin solid
lines with points), bisecting sets (dashed lines) and induced complementary
cells.}
\label{fig:skeleton}
\end{center}
\end{figure}

For each color $i$, there is a hyperplane $P_i$ in $\R^n$ consisting of
all vectors with $\nth{i}{th}$ coordinate equal to $1 / 2$. We can easily
arrange the choice of regular neighborhood $\cN$ of $C$ so that 
$\cN_i = P_i \cap \cN$ is a regular neighborhood of $C \cap P_i$ in
$P_i$. (We call $\cN_i$ a \term{bisecting set} as it intersects $C$ along the
`center' of the reduction in the $\nth{i}{th}$ coordinate direction, see
Figure~\ref{fig:skeleton}.) But then
since $C\cap P_i$ is a cubical complex corresponding to the reduction 
$C^i$, by induction on $n$, we can
assert that $\cN_i$ is an $(n-1)$-ball. Similarly the intersections
$\cN_i \cap \cN_j$ can be arranged to  be regular neighborhoods of
$(d-2)$-maximum classes and are also balls of dimension
$n-2$, etc. In this way, we see that if we can show that $\cN$ is an
$n$-ball, then the induction step will be satisfied and we will have
produced a PL hyperplane arrangement (the system of $\cN_i$ in $\cN$) in a ball. 

\subsubsection{Shifting}

To complete the induction step, we use the technique of shifting~\citep{Al83,Fr83,H95}. In our situation, this can be viewed as the
converse of lifting. Namely if a color $i$ is chosen, then the cubical
complex $C$ has a lifted reduction $C'$ consisting of all $d$-cubes
containing the $\nth{i}{th}$ color.  By shifting, we can move down any of the
lifted components, obtained by splitting $C$ open along $C'$, from the
level $x_i=1$ to the level $x_i=0$, to form a new cubical complex $C^\star$.
We claim that the regular neighborhood of $C$ is a ball if and only if
the same is true for $C^\star$. But this is quite straightforward, since
the operation of shifting can be thought of as sliding components of $C$,
split open along $C'$, continuously from level $x_i=1$ to
$x_i=0$. So there is an isotopy of the attaching maps of  the components
onto the lifted reduction, using the product structure of the latter. It
is easy then to check that this does not affect the homeomorphism type of
the regular neighborhood and so the claim of shift invariance is proved. 

But repeated shifting finishes with the downwards closed maximum class
consisting of all vertices in the $n$-cube with at most $d$ coordinates
being one and the remaining coordinates all being zero. It is easy to see
that the corresponding cubical complex $\tilde C$ is star-like, i.e., contains
all the straight line segments from the origin to any point in $\tilde C$. If
we choose a regular neighborhood $\tilde \cN$ to also be star-like, then
it is obvious that $\tilde \cN$ is an $n$-ball, using radial projection.  Hence our induction is
complete and we have shown that any $d$-maximum class in $\{0,1\}^n$ can be
represented by a family of PL hyperplanes in the $n$-ball.

\subsubsection{Ideal Boundary}

\begin{figure}[t]
\begin{center}
\begin{minipage}[t]{.47\textwidth}
\centering
\includegraphics[width=0.65\textwidth]{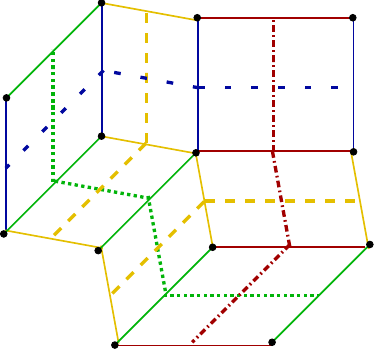}
\caption{The top of Figure~\ref{fig:hyperbolic-ncubes}.(b) (i.e., the $2$-cubes seen from 
above) gives part of
the boundary of a regular neighborhood in $\R^3$.}
\label{fig:hyperbolic4-top}
\end{minipage}\hspace{1.3em}
\begin{minipage}[t]{.47\textwidth}
\centering
\includegraphics[width=0.65\textwidth]{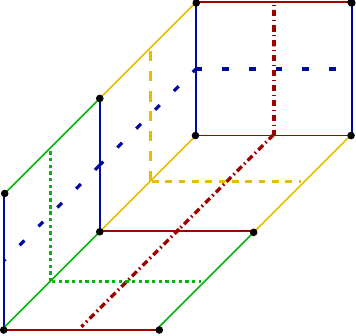}
\caption{The bottom of Figure~\ref{fig:hyperbolic-ncubes}.(b) (i.e., the $2$-cubes seen 
from below) gives the
rest of the boundary of a regular neighborhood.}
\label{fig:hyperbolic4-bottom}
\end{minipage}
\end{center}
\end{figure}

To complete the proof of Theorem~\ref{thm:maximum-is-PL}, let
$\partial \cN =S^{n-1}$ denote the boundary of the $n$-ball $\cN$
constructed above (cf.~Figures~\ref{fig:hyperbolic4-top}
and~\ref{fig:hyperbolic4-bottom}). Each PL hyperplane intersects
this sphere in a PL hypersphere of dimension $n-2$. It remains to show this
arrangement of hyperspheres gives the same cubical complex as $C$, unless
$n=d+1$. 

Suppose that $n>d+1$. Then it is easy to see that each cell
$c$ in the complement of the PL hyperplane arrangement in $\cN$ has part
of its boundary on the ideal boundary $\partial \cN$. Let
$\partial c = \partial c_+\cup \partial c_-$, where $\partial c_+$ is the
intersection of $c$ with the ideal boundary and $\partial c_-$ is the closure
of $ \partial c \setminus \partial c_+$. 

It is now straightforward to verify that the face structure of
$\partial c_+$ is equivalent to the face structure of $\partial c_-$.
Note that any family of at most $d$ PL hyperplanes meet in a PL ball properly embedded
in $\cN$. Since $n>d+1$, the smallest dimension of such a ball is two, and hence its boundary is connected. Then $\partial c_-$ has faces which are PL balls obtained in this way of dimension varying between $n-d$ and $n-1$. Each of these faces has boundary a PL sphere which is a face of 
$\partial c_+$. So this establishes a bijection between the faces of  
$\partial c_+$  and those of $\partial c_-$. It is easy to check that
the cubical complexes corresponding  to the PL hyperplanes and to the PL hyperspheres
are the same. 

Note that if $n=d+1$, then any $d$-maximum
class $C\subseteq\{0,1\}^{d+1}$ is obtained by taking all the $d$-faces of the $(d+1)$-cube
which contain a particular vertex. So $C$ is a $d$-ball and the ideal
boundary of $\cal N$ is a $d$-sphere. The cubical complex associated with
the ideal boundary is the double $2C$ of $C$, i.e., two
copies of $C$ glued together along their boundaries. The proof of
Theorem~\ref{thm:maximum-is-PL} is now complete.

\begin{example}
Consider the bounded below $2$-maximum class $\tilde C\subseteq\{0,1\}^5$. We claim that
$\tilde C$ cannot be realized as an arrangement of PL hyperplanes in the $3$-ball
$B^3$. Note that our method gives $\tilde C$ as an arrangement in $B^5$ and this
example shows that $B^4$ is the best one might hope for in terms of dimension
of the hyperplane arrangement. 

For suppose that $\tilde C$ could be realized by any PL hyperplane
arrangement in $B^3$. Then clearly we can also embed $\tilde C$ into $B^3$.
The vertex $v_0=\{0\}^5$ has link given by the complete graph $K$
on $5$ vertices in $\tilde C$. (By \term{link}, we mean the intersection of the
boundary of a small ball in $B^3$ centered at $v_0$ with $\tilde C$.) But as
is well known, $K$ is not planar, i.e., cannot be embedded into the
plane or $2$-sphere. This contradiction shows that no such arrangement
is possible. 
\end{example}

\subsection{Maximum Classes with Manifold Cubical Complexes}

We prove a partial converse to Corollary~\ref{cor:simple-is-ball}: 
if a $d$-maximum class has a ball as cubical complex, then it can always
be realized by a simple PL hyperplane arrangement in $\R^d$.

\begin{theorem}\label{thm: ball-is-manifold} 
Suppose that $C\subseteq\{0,1\}^n$ is a $d$-maximum class. Then the following
properties of $C$, considered as a cubical complex, are equivalent:

\begin{enumerate}[(i)]

\item There is a  simple arrangement $A$ of $n$ PL hyperplanes in $\R^d$ 
which represents $C$.

\item $C$ is homeomorphic to the $d$-ball.

\item $C$ is a $d$-manifold with boundary. 
\end{enumerate}
\end{theorem}

\begin{proof} To prove (i) implies (ii), we can use exactly the
same argument as Corollary~\ref{cor:simple-is-ball}. Next
(ii) trivially implies (iii). So it remains to show that (iii) implies 
(i). The proof proceeds by double induction on $n,d$.
The initial cases where either $d=1$ or $n=1$ are very easy.

Assume that $C$ is a manifold. Let $p$ denote the $\nth{i}{th}$ coordinate projection.
Then $p(C)$ is obtained by collapsing
$C^i\times [0,1]$ onto $C^i$, where $C^i$ is the reduction. As before, let 
$P_i$ be the linear hyperplane in $\R^n$, where the $\nth{i}{th}$ coordinate
takes value $1/2$. Viewing $C$ as a manifold embedded in the $n$-cube,
since $P_i$ intersects $C$ transversely, we see that $C^i \times \{1 / 2\}$ is
a proper submanifold of $C$. But it is easy to check that collapsing
$C^i\times [0,1]$ to $C^i$ in $C$ produces a new manifold which is again
homeomorphic to $C$. (The product region $C^i\times [0,1]$ in $C$ can be expanded to a larger product region $C^i\times [-\epsilon,1+\epsilon]$ and so collapsing shrinks the larger region to one of the same homeomorphism type, namely $C^i\times [-\epsilon,\epsilon]$ ). So we conclude that the projection $p(C)$ is also a
manifold. By induction on $n$, it follows that there is a PL hyperplane
arrangement $A$, consisting of $n-1$ PL hyperplanes in $B^d$, which
represents $p(C)$.

Next, observe that the reduction $C^i$ can be viewed as a properly embedded
submanifold $M$ in $B^d$, where $M$ is a union of some of the ${(d-1)}$-dimensional
faces of the Voronoi cell decomposition corresponding to $A$, described in
Corollary~\ref{cor:simple-is-ball}. By induction on $d$, we conclude that $C^i$ is also
represented by $n$ PL hyperplanes in $B^{d-1}$. But then since condition (i) implies
(ii), it follows that $M$ is PL homeomorphic to $B^{d-1}$, since the 
underlying cubical complex for $C^i$ is a $(d-1)$-ball. So it follows that 
$A \cup \{M\}$ is a PL hyperplane arrangement in $B^d$ representing $C$. This completes
the proof that condition (iii) implies (i). 
\end{proof}

\section{Corner-Peeling $2$-Maximum Classes}

We give a separate treatment for the case of $2$-maximum classes, since it is simpler than the general case 
and shows by a direct geometric argument, that representation by a simple family of PL hyperplanes or PL hyperspheres
implies a corner-peeling scheme. 

\begin{theorem}\label{thm:peeling-VC-2maximum} 
Every $2$-maximum class can be corner-peeled. 
\end{theorem}

\begin{proof}
By Theorem~\ref{thm:maximum-is-PL}, we can represent any $2$-maximum class
$C\subseteq\{0,1\}^n$ by a simple family of PL hyperspheres $\{S_i\}$ in
$S^{n-1}$.
Every pair of hyperspheres $S_i, S_j$ intersects in an $(n-3)$-sphere $S_{ij}$
and there are no intersection points between any three of
these hyperspheres. Consider the family of spheres $S_{ij}$, for $i$ fixed.
These are disjoint hyperspheres in $S_i$ so we can choose an innermost
one $S_{ik}$ which bounds an ${(n-2)}$-ball $B_1$ in $S_i$ not containing
any other of these spheres. Moreover there are two balls $B_2, B_3$ bounded
by $S_{ik}$ on $S_k$. We call the two $(n-1)$-balls  $Q_2,Q_3$ bounded
by $B_1 \cup B_2$, $B_1 \cup B_3$ respectively in $S^{n-1}$, which
intersect only along $B_1$, $\it quadrants$. 

Assume $B_2$ is innermost on $S_k$. Then the quadrant 
$Q_2$ has both faces $B_1,B_2$ innermost. It is easy to see that such
a quadrant corresponds to a corner vertex
in $C$ which can be peeled. Moreover, after peeling, we still have a
family of PL hyperspheres which give an arrangement corresponding to the
new peeled class. The only difference is that cell $Q_2$ disappears, by
interchanging $B_1,B_2$ on the corresponding spheres $S_i, S_k$ and then
slightly pulling the faces apart. (If $n=3$, we can visualize a pair of disks on two intersecting spheres with a common boundary circle. Then peeling can be viewed as moving these two disks until they coincide and then pulling the first past the second). So it is clear that if we can repeatedly show that
a quadrant can be found with two innermost faces, until all the intersections
between the hyperspheres have been removed, then we will have corner-peeled
$C$ to a $1$-maximum class, i.e., a tree. So peeling
will be established. 

Suppose neither of the two quadrants $Q_2,Q_3$ has both faces innermost.
Consider $Q_2$ say and let $\{S_\alpha\}$ be the family of spheres
intersecting the interior of the face $B_2$. Amongst these spheres, there is
clearly at least one $S_\beta$ so that the intersection $S_{k\beta}$ is
innermost on $S_k$. But then $S_{k\beta}$ bounds an innermost ball $B_4$ in
$S_k$ whose interior is disjoint from all the spheres $\{S_\alpha\}$.
Similarly, we see that $S_{k\beta}$ bounds a ball $B_5$ which is the
intersection of the sphere $S_\beta$ with the quadrant $Q_2$. We get a new
quadrant bounded by $B_4 \cup B_5$ which is strictly smaller than $Q_2$ and
has at least one innermost face. But clearly this process must terminate---we
cannot keep finding smaller and smaller quadrants and so a smallest one must
have both faces innermost.
\end{proof}

\section{Corner-Peeling Finite Maximum Classes}

Above, simple PL hyperplane arrangements in the $n$-ball $B^n$ are
defined. For the purposes of this section, we study a slightly more general
class of arrangements. 

\begin{definition}\label{defn:contractible}
Suppose that a finite arrangement $\cP$ of PL hyperplanes
$\{P_\alpha\}$, each properly embedded in an $n$-ball $B^n$,
satisfies the following conditions:
\begin{enumerate}[i.]
\item Each $k$-subcollection of hyperplanes either intersects transversely in a
PL plane of dimension $n-k$, or has an empty intersection; and
\item The maximum number of hyperplanes which mutually intersect is $d \le n$. 
\end{enumerate}
Then we say that the arrangement $\cP$ is \term{$d$-contractible}.
\end{definition}

The arrangements  in Definition~\ref{defn:contractible} are called $d$-contractible because
we prove later that their corresponding one-inclusion graphs are strongly contractible cubical complexes of dimension $d$.
Moreover we now prove that the corresponding one-inclusion graphs have VC dimension exactly $d$.

\begin{lemma}
The one-inclusion graph $\Gamma$ corresponding to a $d$-contractible arrangement $\cP$ has VC dimension $d$.
\end{lemma}

\begin{proof}
We observe first of all, that since $\cP$ has a subcollection of $d$ hyperplanes which mutually intersect, the corresponding one-inclusion graph $\Gamma$ has a $d$-subcube, when considered as a cubical complex. But then the VC dimension of $\Gamma$ is clearly at least $d$. On the other hand, suppose that the VC dimension of $\Gamma$ was greater than $d$. Then there is a projection of $\Gamma$ which shatters some $(d+1)$-cube. But this projection can be viewed as deleting all the hyperplanes of $\cP$ except for a subcollection of $d+1$ hyperplanes. However, by assumption, such a collection cannot have any mutual intersection points. It is easy to see that any such an arrangement has at most $2^{d+1}-1$ complementary regions and hence cannot represent the $(d+1)$-cube. This completes the proof. 
\end{proof}

\begin{definition}
A one-inclusion graph $\Gamma$ is \term{strongly contractible} if it is
contractible as a cubical complex and moreover, all reductions and multiple
reductions of $\Gamma$ are also contractible. 
\end{definition}

\begin{definition}
The \term{complexity} of a PL hyperplane arrangement $\cP$ is the
lexicographically ordered pair $(r,s)$, where $r$ is the number of regions in
the complement of $\cP$, and $s$ is the smallest number of regions
in any half space on one side of an individual hyperplane in $\cP$. 
\end{definition}

We allow several different hyperplanes to be used for a single
sweeping process. So a hyperplane $P$ may start sweeping
across an arrangement $\cP$. One of the half spaces defined by $P$ can become a
new ball $B_+$ with a new arrangement $\cP_+$ defined by restriction of $\cP$
to the half space $B_+$. Then a second generic hyperplane $P^\prime$ can start
sweeping across this new arrangement $\cP_+$. This process may occur several
times. It is easy to see that sweeping a single generic 
hyperplane as in Theorem~\ref{thm:sweeping-hyperbolic-is-corner}, applies to such a multi-hyperplane process. Below we
show that a suitable multiple sweeping of a PL hyperplane arrangement $\cP$
gives a corner-peeling sequence of all finite maximum classes.

The following states our main theorem.

\begin{theorem}\label{thm:peel-finite}
Assume that $\cP$ is a $d$-contractible PL hyperplane arrangement in the $n$-ball $B^n$. 
Then there is a $d$-corner-peeling scheme for this collection $\cP$.
\end{theorem}

\begin{corollary}
There is no constant $k$ so that every finite maximal class of VC
dimension $d$ can be embedded into a maximum class of VC dimension $d+k$.
\end{corollary}

\begin{proof}
 By Theorem~\ref{thm:peel-finite}, every maximum class has a peeling scheme
which successively removes vertices from the one-inclusion graph, so that
the vertices being discarded never have degree more than $d$. But~\citet{RBR07a}gave examples of maximal classes of VC dimension $d$ which have a core of
the one-inclusion graph of size $d+k$ for any constant $k$. Recall that
a core is a subgraph and its size is the minimum degree of all the
vertices. Having a peeling scheme gives an upper bound on the size of
any core and so the result follows. 
\end{proof}

\subsection{Pachner Moves}

\citet{P87} showed that triangulations of manifolds which are
combinatorially equivalent after subdivision are also equivalent by a
series of moves which are now referred to as Pachner moves. 
For the main result of this paper, we need a version of Pachner moves for
cubical structures rather than simplicial ones. The main idea of Pachner moves
remain the same. 

\begin{figure}[t]
\begin{center}
\begin{minipage}[t]{0.47\textwidth}
\centering
\includegraphics[width=0.8\textwidth]{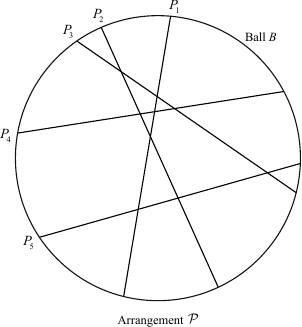}
\caption{An example piecewise-linear hyperplane arrangement $\cP$.}
\label{fig:splitting-and-pachner-a}
\end{minipage}\hspace{1em}
\begin{minipage}[t]{0.47\textwidth}
\centering
\includegraphics[width=0.8\textwidth]{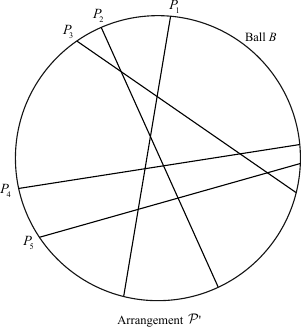}
\caption{Result of a Pachner move of hyperplane $P_4$ on
$\cP$ in Figure~\ref{fig:splitting-and-pachner-a}.}
\label{fig:splitting-and-pachner-b}
\end{minipage}
\end{center}
\end{figure}

A \term{Pachner move} replaces a topological $d$-ball $U$ divided
into $d$-cubes, with
another ball $U^\prime$ with the same $(d-1)$-cubical boundary but with a
different interior cubical structure. In dimension $d=2$, for example, such an
initial ball $U$ can be constructed by taking three $2$-cubes forming a
hexagonal disk and
in dimension $d=3$, four $3$-cubes form a rhombic dodecahedron, which is a
polyhedron $U$ with $12$ quadrilateral faces in its boundary. The set $U^\prime$
of $d$-cubes is attached to the same boundary as for $U$, i.e.,
$\partial U = \partial U^\prime$, as cubical complexes homeomorphic to the
$(d-1)$-sphere. Moreover, $U^\prime$ and $U$ are isomorphic cubical complexes, but the gluing between their boundaries produces the boundary of the 
$3$- or $4$-cube, as a $2$- or $3$- dimensional cubical structure on the $2$- or $3$-sphere respectively. 

To better understand this move, consider the
cubical face structure of the boundary $V$ of the $(d+1)$-cube. This is a
$d$-sphere containing $2d+2$ cubes, each of dimension $d$. There are many
embeddings of the $(d-1)$-sphere as a cubical subcomplex into $V$, dividing it into a pair of $d$-balls.
One ball is combinatorially identical to  $U$ and the other to $U^\prime$.

There are a whole series of Pachner moves in each dimension $d$, but we are
only interested in the ones where the pair of balls $U, U^\prime$ have the same
numbers of $d$-cubes. In Figures~\ref{fig:splitting-and-pachner-a}
and~\ref{fig:splitting-and-pachner-b} a change in a
hyperplane arrangement is shown, which corresponds to a Pachner move on the
corresponding one-inclusion graph (considered as a cubical complex). 

\subsection{Proof of Main Theorem}

\begin{figure}[t]
\begin{center}
\centering
\includegraphics[width=0.5\textwidth]{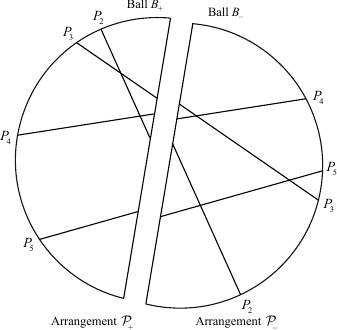}\label{fig:splitting-and-pachner-c}
\caption{Result of splitting $\cP$ in Figure~\ref{fig:splitting-and-pachner-a}
along hyperplane $P_1$.}
\end{center}
\end{figure}

The proof is by induction on the  complexity of $\cP$. 
Since we are dealing with the class of $d$-contractible PL hyperplane
arrangements, it is easy to see that if any such $\cP$ is
\term{split open} along some fixed hyperplane $P_1$ in the
arrangement (see Figure~\ref{fig:splitting-and-pachner-c}),
then the result is two new arrangements $\cP_+, \cP_-$ each of which
contains fewer hyperplanes and also fewer complementary regions than
the initial one. The new arrangements have smaller complexity than
$\cP$ and are $k-, k^\prime$-contractible for some $k, k^\prime \le d$. 
This is the key idea of the construction. 

To examine this splitting process in detail, first note that each
hyperplane $P_\alpha$ of $\cP$ is either disjoint from $P_1$ or splits
along $P_1$ into two hyperplanes $P_\alpha^+, P_\alpha^-$. We can now
construct the new PL hyperplane arrangements $\cP_+, \cP_-$ in the balls
$B_+, B_-$ obtained by splitting $B$ along $P_1$. Note that
$\partial B_+ = P_1 \cup D_+$ and $\partial B_-=P_1 \cup D_-$ where
$D_+, D_-$ are balls of dimension $n-1$ which have a common boundary
with $P_1$. 
It is easy to verify that $\cP_+, \cP_-$ satisfy similar hypotheses to
the original arrangement. Observe that the maximum number of mutually
intersecting hyperplanes in $\cP_+, \cP_-$ may decrease relative to
this number for $\cP$, after the splitting operation. The reason is that
the hyperplane $P_1$ `disappears' after splitting and so if all maximum
subcollections of $\cP$ which mutually intersect, all contain $P_1$, then this number 
is smaller for $\cP_+, \cP_-$ as compared to the initial arrangement $\cP$. 
This number shows that  $\cP_+, \cP_-$ can be $k$- or $k^\prime$-contractible, for 
$k, k^\prime < d$ as well as the cases where $k, k^\prime = d$. 

Start the induction with any arrangement with one hyperplane. This gives two regions and complexity $(2,1)$.
The corresponding graph has one edge and two vertices and obviously can be corner-peeled.

We now describe the inductive step. There are two cases. In the first, assume the arrangement has
complexity $(r,1)$. The corresponding graph has a vertex which belongs to only
one edge, so can be corner-peeled. This gives an arrangement with fewer hyperplanes and clearly the complexity has decreased to $(r-1,s)$
for some $s$. This completes the inductive step for the first case.

For the second case,  assume that all $d$-contractible
hyperplane arrangements with complexity smaller than $(r,s)$ have
corner-peeling sequences and $s>1$.  Choose any $d$-contractible hyperplane arrangement
$\cP$ with complexity $(r,s)$. Select a hyperplane $P_1$ which splits the
arrangement into two smaller arrangements $\cP_+, \cP_-$ in the balls
$B_+, B_-$. By our definition of complexity, it is easy to see that however we choose $P_1$, the complexity of each of $\cP_+,\cP_-$  
will be less than that of $\cP$. However, a key requirement for the proof will be that we select $P_1$ so that it has precisely $s$ complementary regions for $\cP_+$,
i.e., $P_1$ has fewest complementary regions in one of its halfspaces, amongst all hyperplanes in the arrangement. 

\begin{figure}[t]
\begin{center}
\centering
\includegraphics[width=\textwidth]{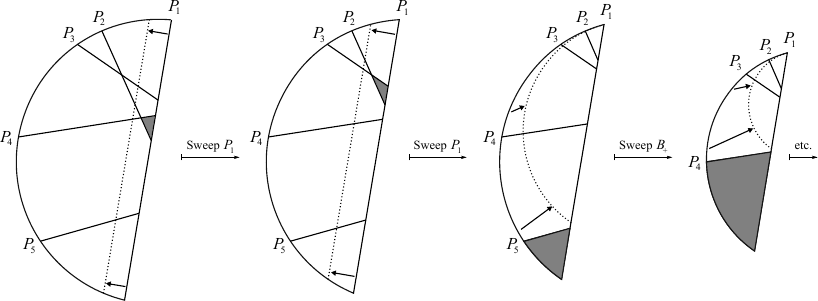}
\caption{Partial corner-peeling sequence for the $(B_+,P_+)$ arrangement split from the arrangement of Figure~\ref{fig:splitting-and-pachner-c}, in the proof of
Theorem~\ref{thm:peel-finite}.}
\label{fig:corner-peeling}
\end{center}
\end{figure}

Since $\cP_+$  has smaller complexity than $(r,s)$, by our inductive
hypothesis, it can be corner-peeled (cf.~Figure~\ref{fig:corner-peeling}). If
any of the corner-peeling moves of $\cP_+$ is a
corner-peeling move for $\cP$, then the argument follows. For any
corner-peeling move of $\cP$ gives a PL hyperplane arrangement with fewer
complementary cells than $\cP$ and thus smaller complexity than $(r,s)$.
Hence by the inductive hypothesis, it follows that $\cP$ can be corner-peeled. 

Next, suppose that no corner-peeling move of $\cP_+$ is a
corner-peeling move for $\cP$. In particular, the first corner-peeling move
for $\cP_+$ must occur for a cell $R_+$ in the complement
of $\cP$, which is adjacent to $P_1$. (Clearly any
corner-peeling move for $\cP_+$, which occurs at a region
$R_1$ with a face on $D_+$, will be a corner-peeling
move for $\cP$.) $R_+$ must be a product of a
$d^\prime$-simplex $\Delta$ with a copy of ${\bR}^{n-d^\prime}$, with one face
on $P_1$ and the other faces on planes of $\cP$. This is because a
corner-peeling move can only occur at a cell with this type of face
structure, as described in Theorem~\ref{thm:sweeping-hyperbolic-is-corner}. The corresponding effect on the
one-inclusion graph is peeling of a vertex which is a corner of a $d^\prime$-cube 
in the binary class corresponding to the arrangement $\cP_+$,
where $d^\prime \le d$.

Now even though such a cell $R_+$ does not give a corner-peeling
move for $\cP$, we can push $P_1$ across $R_+$. The
effect of this is to move the complementary cell $R_+$ from $B^+$ to
 $B^-$. Moreover, since we assumed that the hyperplane $P_1$
satisfies $\cP^+$ has a minimum number $s$ of
complementary regions,  it follows that the move pushing $P_1$ across $R_+$
 produces a new arrangement
$\cP^*$ with smaller complexity $(r,s-1)$ than the original 
arrangement $\cP$. Hence by our inductive assumption, $\cP^*$
admits a corner-peeling sequence. 

To complete the proof, we need to show that existence of a corner-peeling 
sequence for $\cP^*$ implies that the original arrangement $\cP$ has at
least one corner-peeling move. Recall that $R_+$ has face structure given by
$\Delta \times {\bR}^{n-d^\prime}$, with one face on $P_1$ and the other faces
on planes of $\cP$. Consider the subcomplex $U$ of the one-inclusion graph
consisting of all the regions sharing a vertex or face of dimension $k$ for
$1 \le k \le n-1$ with $R_+$. It is not difficult to see that $U$ is a
$d^\prime$-ball consisting of $d^\prime+1$ cubes, each of dimension $d^\prime$. 
(As examples, if $d^\prime=2$, $U$ consists of $3$ $2$-cubes forming a hexagon and if 
$d^\prime=3$, $U$ consists of $4$ $3$-cubes with boundary a rhombic dodecahedron.)

\begin{figure}[t]
\begin{center}
\centering
\includegraphics[width=\textwidth]{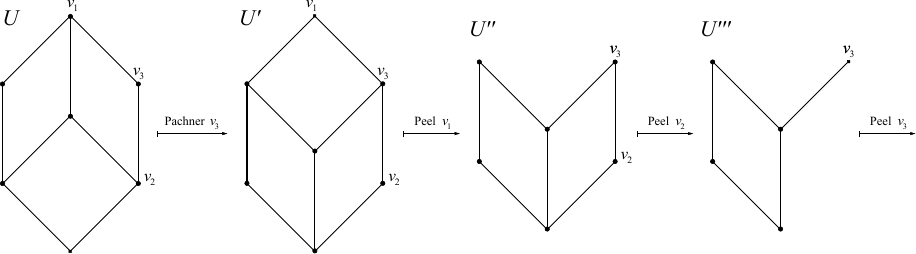}
\caption{A 2-maximum complex in the 3-cube.  After a Pachner move vertices
$v_1, v_2, v_3$, etc. can be corner-peeled.}
\label{fig:corner-peeling-pachner}
\end{center}
\end{figure}

Consider the first corner-peeling move on the arrangement $\cP^*$. Note that
the one-inclusion graphs of $\cP^*$  and $\cP$  differ precisely by replacing
$U$ with $U^\prime$, i.e., by a Pachner move.  Hence this first corner-peeling
move must occur at a vertex $v_1$ whose degree 
is affected by this replacement, since otherwise, the corner-peeling move would also apply to $\cP$  and the
proof would be complete. In fact, if $v_1$ has the same number of adjacent edges before and after the Pachner move, 
then it must belong to the same single maximum
dimension cube before and after the Pachner move. (The only cubes altered by the Pachner move are the ones in $U$.)
It is easy to see that, $v_1$ must belong to 
$\partial U=\partial U^\prime$ and must have degree $d^\prime$ in $\cP^*$. So $v_1$ is a corner
of a single $d^\prime$-cube for $U^\prime$ and does not belong to any other edges or
cubes of the one inclusion graph for $\cP^*$. In $U$ (and hence also in $\cP$),
$v_1$ belongs to $d^\prime$-cubes of dimension $d^\prime$ and so has degree
$d^\prime+1$. After peeling away $v_1$ and its corresponding $d^\prime$-cube,
we still have a $d^\prime$-ball with only $d^\prime$-cubes, (cf.~Figure~\ref{fig:corner-peeling-pachner}).

Consider the next corner-peeling move. We claim that it must again be at a
vertex $v_2$ belonging to 
$\partial U^\prime$. The reason is that only vertices belonging to $U^\prime$
 have degree reduced by our first corner-peeling move. So if this second move
did not occur at a vertex of $U^\prime$, then it could be used as a
corner-peeling move of our initial arrangement $\cP$. There may be several
choices for $v_2$. For example, if $d^\prime=2$, then $U^\prime$ is a
hexagonal disk and removing one $2$-cube from $U^\prime$ gives a choice which
could be either of the two vertices which are corners of a single $2$-cube in
$U^\prime$, (cf.~Figure~\ref{fig:corner-peeling-pachner}). 
Note that a vertex which is a corner of a single cube
in $U^\prime$ remains so after corner-peeling at $v_1$. Note also that
$v_2$ cannot belong to any edges of the one-inclusion graph which are not 
in $U^\prime$, as for $v_1$, if $v_2$ can be used for corner-peeling.  

We can continue examining corner-peeling moves of $\cP^*$
and find that all must occur at vertices in $\partial U^\prime$, until the unique
interior vertex is ready to be peeled, i.e., belongs to a single cube. (See
Figure~\ref{fig:corner-peeling-pachner}.) The key to understanding this is that firstly,
when we initially peel only vertices in $\partial U^\prime$, these are not adjacent to any vertices
of the one-inclusion graph outside $ U^\prime$ and so cannot produce any new opportunities
for corner-peeling of vertices not in $U^\prime$. Secondly, if the unique interior vertex $v$ of $U^\prime$ 
can be corner-peeled, after sufficiently many vertices in 
$\partial U^\prime$ have been peeled, then new vertices in $\partial U^\prime$
become candidates for peeling. For although these latter vertices may be adjacent to vertices outside $U^\prime$, 
after $v$ has been peeled, they may become a corner vertex of a unique maximal cube. 

But now a final careful examination of this situation shows that 
there must be at least one vertex of $U$ which belongs to a single
$d^\prime$-cube in $U$ and to no other edges in $\cP$. So this will give our
initial corner-peeling move of $\cP$. 

To elaborate, we can describe $U$ as the set of $d^\prime$-cubes which share the vertex $(0,0, \dots ,0)$
in the $(d^\prime +1)$-cube $\{0,1\}^{d^\prime +1}$. Then $U^\prime$ consists of all the
$d^\prime$-cubes in $\{0,1\}^{d^\prime +1}$ which contain the vertex $(1,1, \dots ,1)$.
Now assume that an initial sequence of corner-peeling of vertices in $\partial U^\prime$ allows the next step to 
be corner-peeling of the unique interior vertex $v$. Note that in the notation above, $v$ corresponds to the vertex $(1,1, \dots ,1)$.

As in Figure~\ref{fig:corner-peeling-pachner}, we may assume that after the corner-peeling corresponding to the initial sequence of vertices in $\partial U^\prime$,
that there is a single $d^\prime$-cube left in $U^\prime$ containing $v$. Without loss of generality, suppose this is the cube with vertices with $x_1=1$
where the coordinates are $x_1,x_2, \dots, x_{d^\prime +1}$ in $\{0,1\}^{d^\prime +1}$. But then, it follows that there are no vertices outside $U^\prime$ adjacent to 
any of the initial sequence of vertices, which are all the vertices in $\{0,1\}^{d^\prime +1}$ with $x_1=0$, except for $(0,0, \dots ,0)$. But now the vertex $(0,1,\dots, 1)$ 
has the property that we want - it is contained in a unique $d^\prime$-cube in $U$ and is adjacent to no other vertices outside $U$. This completes the proof. 

\subsection{Peeling Classes with Generic Linear or Generic Hyperbolic Arrangements}
\label{subsec:generic-arrangements}

\begin{definition}
A linear or hyperbolic hyperplane arrangement $\cP$ in $\R^n$ or $\bbH^n$ respectively, is called \emph{generic}, if any subcollection of $k$ hyperplanes of $\cP$, for $2 \le k \le n$
has the property that there are no intersection points or the subcollection intersects transversely in a plane of dimension $n-k$. 
\end{definition}

\begin{corollary}\label{cor:linear}
Suppose $\cP$ is a generic linear or hyperbolic hyperplane arrangement in $\R^n$ or $\bbH^n$ and amongst all subcollections of $\cP$, the largest with an intersection point in common, has $d$ hyperplanes. Then $\cP$ admits a $d$-corner-peeling scheme.
\end{corollary}

\begin{remark}
The proof of Corollary~\ref{cor:linear} is immediate since it is obvious that any generic linear or hyperbolic hyperplane arrangement is a $d$-contractible PL hyperplane arrangement, where
$d$ is the cardinality of the largest subcollection of hyperplanes which mutually intersect. 
Note that many generic linear, hyperbolic or $d$-contractible PL hyperplane arrangements do not embed in any simple linear, hyperbolic or PL hyperplane arrangement. For if there are two hyperplanes in $\cP$ which are disjoint, then this is an obstruction to enlarging the arrangement by adding additional hyperplanes to obtain a simple arrangement. Hence this shows that
Theorem~\ref{thm:peel-finite}  produces compression schemes, by corner-peeling, for a considerably larger class of one-inclusion graphs than just maximum one-inclusion graphs. However it seems possible that $d$-contractible PL hyperplanes always embed in $d$-maximum classes, by `undoing' the operation of sweeping and corner-peeling, which pulls apart the hyperplanes. 
\end{remark}

\section{Peeling Infinite Maximum Classes with Finite Dimensional Arrangements}

We seek infinite classes represented by arrangements satisfying
the same conditions as above. Note that any finite
subclass of such an infinite class then satisfies these conditions and
so can be corner-peeled. Hence any such a finite subclass has a
complementary region $R$ which has face structure of the product of a
$d^\prime$-simplex with a copy of ${\bR}^{n-d^\prime}$ with one face
on the boundary of $B^n$. To find such a region in the complement of
our infinite collection $\cP$, we must impose some conditions. 

One convenient condition (cf the proof of Theorem~\ref{thm:peel-finite}) is that a hyperplane
$P_\alpha$ in $\cP$ can be found which splits $B^n$ into
pieces $B_+, B_-$ so that one, say $B_+$ gives a new arrangement for
which the maximum number of mutually intersecting hyperplanes is strictly
less than that for $\cP$. Assume that the new arrangement satisfies a
similar condition, and we can keep splitting until we get to disjoint hyperplanes. 

It is not hard to prove that such arrangements always have peeling
sequences. Moreover the peeling sequence does give a compression scheme.
This sketch establishes the following.

\begin{theorem}\label{thm:infinite}
Suppose that a countably infinite collection $\cP$ of PL hyperplanes
$\{P_\alpha\}$, each properly embedded in an $n$-ball $B^n$, satisfies
the following conditions:
\begin{enumerate}[i.]
\item $\cP$ satisfies the conditions of $d$-contractible arrangements as in Definition~\ref{defn:contractible}
and
\item There is an ordering of the planes in $\cP$ so that if we split $B^n$
successively along the planes, then at each stage, at least one of the two
resulting balls has an arrangement with a smaller maximum number of planes which
mutually intersect.
\end{enumerate}
Then there is a $d$-corner-peeling scheme for $\cP$, and this
provides a $d$-unlabeled compression scheme. 
\end{theorem} 

\begin{example}
\citet{RR08} give an example that satisfies the assumptions of
Theorem~\ref{thm:infinite}. Namely in $\bR^n$ choose the positive octant
$\cO = \{(x_1,x_2, \dots x_n): x_i \ge 0\}$. Inside $\cO$ choose the
collection of hyperplanes given by $x_i=m$ for all $1 \le i \le n$ and $m \ge 1$
a positive integer. There are many more examples, we present only a very simple
model here. Take a graph inside the unit disk $D$ with a single vertex of
degree $3$ and the three end vertices on $\partial D$. Now choose a collection
of disjoint embedded arcs representing hyperplanes with ends on $\partial D$
and meeting one of the edges of the graph in a single point. We choose finitely
many such arcs along two of the graph edges and an infinite collection along
one arc. This gives a very simple family of hyperplanes satisfying the
hypotheses of Theorem~\ref{thm:infinite}. Higher dimensional examples with
intersecting hyperplanes based on arbitrary trees can be constructed in a
similar manner. 
\end{example}

\section{Contractibility, Peeling and Arrangements}

In this section, we characterize the concept classes which have one-inclusion
graphs representable by $d$-contractible PL hyperplane arrangements.

\begin{theorem}\label{thm:contractible}
Assume that $\cC$ is a concept class in the binary $n$-cube
and $d$ is the largest dimension of embedded cubes in its one-inclusion graph $\Gamma$. 
The following are equivalent.
\begin{enumerate}[i.]
\item $\Gamma$ is a strongly contractible cubical complex. 
\item There is a $d$-contractible PL hyperplane arrangement $\cP$ in an
$n$-ball which represents $\Gamma$.
\end{enumerate}
\end{theorem}

\begin{proof}
To prove that \emph{i} implies \emph{ii}, we use some important ideas in the 
topology of manifolds. The cubical complex $\cC$ is naturally embedded into the binary $n$-cube, which
can be considered as an $n$-ball $B^n$. A regular neighborhood $N$ of $\cC$
homotopy retracts onto $\cC$ and so is contractible. Now we can use a standard
argument from algebraic and geometric topology to prove that $N$ is a ball. Firstly,
$\partial N$ is simply connected, assuming that $n-d>2$. For given a loop in
$\partial N$, it bounds a disk in $N$ by contractibility. Since $\cC$ is a
$d$-dimensional complex and $n-d>2$ it follows that this disk can be pushed off
$\cC$ by transversality and then pushed into $\partial N$. But now we can
follow a standard argument using the solution of the Poincar\'e conjecture in
all dimensions~\citep{P02, F82, S61}. By duality, it follows that $\partial N$
is a homotopy $(n-1)$-sphere and so by the Poincar\'e conjecture, $\partial N$
is an $(n-1)$-sphere. Another application of the Poincar\'e conjecture shows
that $N$ is an $n$-ball. 

Next, the bisecting planes of the binary $n$-cube meet the $n$-ball $N$ in
neighborhoods of reductions. Hence the assumption that each reduction is
contractible enables us to conclude that these intersections are also PL
hyperplanes in $N$. Therefore the PL hyperplane arrangement has been
constructed which represents $\Gamma$. It is easy to see that this arrangement
is indeed $d$-contractible, since strong contractibility implies that all
multiple reductions are contractible and so intersections of subfamilies of PL
hyperplanes are either empty or are contractible and hence planes, by the same
argument as the previous paragraph. (Note that such intersections correspond to multiple reductions of $\Gamma$.)

Finally to show that \emph{ii} implies \emph{i}, by Theorem~\ref{thm:peel-finite}, a
$d$-contractible PL hyperplane arrangement $\cP$ has a peeling sequence and so
the corresponding one-inclusion graph $\Gamma$ is contractible. This follows
since a corner-peeling move can be viewed as a homotopy retraction. But then
reductions and multiple reductions are also represented by
$d^\prime$-contractible hyperplane arrangements, since these correspond to the
restriction of $\cP$ to the intersection of a finite subfamily of hyperplanes
of $\cP$.  It is straightforward to check that these new arrangements are
$d^\prime$-contractible, completing the proof.
\end{proof}

\begin{remark}
Note that any one-inclusion graph $\Gamma$ which satisfies the hypotheses of Theorem~\ref{thm:contractible} admits a corner-peeling sequence. From the proof above, $\Gamma$ must be contractible if it has a peeling sequence.  However $\Gamma$ does not have to be strongly contractible. A simple example can be found in the binary $3$-cube, with coordinate directions $x,y,z$. Define $\Gamma$ to be the union of four edges, labeled $x,y,z,x$. It is easy to see that $\Gamma$ has a peeling sequence and is contractible but not strongly contractible. For the bisecting hyperplane transverse to the $x$ direction meets $\Gamma$ in two points, so the reduction $\Gamma^x$ is a pair of vertices, which is not contractible. 

Note that all maximum classes are strongly contractible, as are also all linear and hyperbolic arrangements, by Corollary~\ref{cor:linear} and Theorem~\ref{thm:contractible}. 

\end{remark}

\section{Future Directions: Compression Schemes for Maximal Classes}

In this section, we compare two maximal classes of VC dimension $2$ in the
binary $4$-cube. For the first, we show that the one-inclusion graph is not
contractible and therefore there is no peeling or corner peeling scheme. There
is an unlabeled compression scheme, but this is not associated with either
peeling or a hyperplane arrangement. For the second, the one-inclusion graph
is contractible but not strongly contractible. However there are simple corner
peeling schemes and a related compression scheme. Note that the relation
between the compression scheme and the corner peeling scheme is not as
straightforward as in our main result above. Finally for the second example,
there is a non simple hyperplane arrangement consisting of lines in the
hyperbolic plane which represents the class. It would be interesting to know
if there are many maximal classes which admit such non simple representations
and if there is a general procedure to find associated compression schemes.

\begin{figure}[!t]
\begin{center}
\begin{minipage}[t]{.48\textwidth}
\centering
\begin{tabular}{cc}
\hline
\textbf{Concept} & \textbf{Label} \\
\hline
0000 & $\emptyset$ \\
1000 & $x_1$ \\
0100 & $x_2$ \\
0010 & $x_3$ \\
0001 & $x_4$ \\
1100 & $x_1 x_2$ \\
0011 & $x_3 x_4$ \\
0110 & $x_2 x_3$ \\
1001 & $x_1 x_4$ \\
1111 &  $x_1 x_3$,  $x_2 x_4$ \\
\hline
\end{tabular}
\caption{Example~\ref{eg:future-1} VC-2 maximal class.}
\label{fig:future-1}
\end{minipage}\hspace{1.0em}
\begin{minipage}[t]{.48\textwidth}
\centering
\begin{tabular}{cc}
\hline
\textbf{Concept} & \textbf{Label} \\
\hline
0000 & $\emptyset$ \\
1000 &  $x_1$ \\
0100 & $x_2$ \\
0010 & $x_3$ \\
1100 & $x_1 x_2$ \\
0110 & $x_2 x_3$ \\
1010 & $x_1 x_3$ \\
1011 &  $x_2 x_4$ \\
1101 & $x_3 x_4$ \\
0111 &  $x_1 x_4$ \\
\hline
\end{tabular}
\caption{Example~\ref{eg:future-2} VC-2 maximal class.}
\label{fig:future-2}
\end{minipage}
\end{center}
\end{figure}

\begin{example}\label{eg:future-1}
Let $\mathcal C$ be the maximal class of VC dimension $2$ in the $4$-cube with
concepts and labels shown in Figure~\ref{fig:future-1}.
This forms an unlabeled compression scheme. Note that the one-inclusion graph
is not connected, consisting of four $2$-cubes with common vertex at the
origin 0000 and an isolated vertex at 1111. So since a contractible complex is
connected, the one-inclusion graph cannot be contractible. Moreover any
hyperplane arrangement represents a connected complex so there cannot be such
an arrangement for this example. This example is the same class (up to
flipping coordinate labels) as in~\citep[Table 2]{KW07} but there appear to be
some
errors there in describing the compression scheme. 
\end{example}

\begin{example}\label{eg:future-2}
Let $\mathcal C$ be the maximal class of VC dimension $2$ in the $4$-cube with
concepts and labels defined in Figure~\ref{fig:future-2}.
The class is enlarged by adding an extra vertex 1111 $x_4$ to complete the
labeling.

This forms an unlabeled compression scheme and is the same as
in~\citep[Table 1]{KW07}. The one-inclusion graph is contractible, consisting of three $2$-cubes
with common vertex 0100 and three edges attached to these $2$-cubes. It is
easy to form a hyperbolic line arrangement consisting of three lines meeting
in three points forming a triangle and three further lines near the boundary
of the hyperbolic plane which do not meet any other line. 

It is easy to see that there is a corner peeling sequence, but there is not
such an obvious way of using this to form a compression scheme. The idea is
that the label $x_1 x_4$ comes from picking the origin at 0000 and considering
the shortest path to the origin as giving the label. There are numerous ways
of corner peeling this one-inclusion complex. The only other comment is that
the final vertices 0111, 1011, 1101 and 1111 are labeled in a different manner. 
Namely putting the origin at 0000 means that 0111 has shortest path with label
$x_2 x_3 x_4$. We replace this by the label $x_1 x_4$ since clearly this
satisfies the no-clashing condition. Then the final vertex 1111 has the
remaining label $x_4$ to uniquely specify it.
\end{example}

\section{Conclusions and Open Problems}

We saw in Corollary~\ref{cor:simple-is-ball} that $d$-maximum
classes represented by simple linear hyperplane arrangements in
$\R^d$ have underlying cubical complexes that are
homeomorphic to a $d$-ball. Hence the VC dimension and the dimension
of the cubical complex are the same. Moreover in
Theorem~\ref{thm: ball-is-manifold}, we proved that $d$-maximum classes
represented by PL hyperplane arrangements in $\R^d$ are those whose
underlying cubical complexes are manifolds or equivalently $d$-balls.

\begin{question}
Does every simple PL hyperplane arrangement in $B^d$, where every subcollection
of $d$ planes transversely meet in a point, represent the same concept class
as some simple linear hyperplane arrangement?
\end{question}
 
\begin{question}
What is the connection between the VC dimension of a maximum class induced
by a simple hyperbolic hyperplane arrangement and the smallest dimension
of hyperbolic space containing such an arrangement? In particular, can
the hyperbolic space dimension be chosen to only depend on the VC dimension
and not the dimension of the binary cube containing the class?
\end{question}

We gave an example of a $2$-maximum class in the $5$-cube that
cannot be
realized as a hyperbolic hyperplane arrangement in $\H^3$. Note that
the Whitney embedding theorem~\citep{RS82} proves that any
cubical complex of dimension $d$ embeds in $\R^{2d}$. Can
such an embedding be used to construct a hyperbolic arrangement in
$\H^{2d}$ or a PL arrangement in $\R^{2d}$?

The structure of the boundary of a maximum class is strongly related to
corner-peeling. For Euclidean hyperplane arrangements, the boundary of the
corresponding maximum class is homeomorphic to a sphere by
Corollaries~\ref{cor:generic-intersect-is-submax} and~\ref{cor:simple-is-ball}.

\begin{question}
Is there a characterization of the 
cubical complexes that can occur as the boundary of a maximum class? 
Characterize maximum classes with isomorphic boundaries.
\end{question}

\begin{question}
Does a corner-peeling scheme exist with corner vertex sequence
having minimum degree?
\end{question}

Theorem~\ref{thm:maximum-is-PL} suggests the following.

\begin{question}
Can any $d$-maximum class in $\{0,1\}^n$ be represented by a simple
arrangement of hyperplanes in $\bbH^n$? 
\end{question}

\begin{question} Which compression schemes arise from sweeping across
simple hyperbolic hyperplane arrangements?
\end{question}

\citet{KW07} note that there are unlabeled compression schemes that are cyclic.
In Proposition~\ref{prop:acyclic} we show that corner-peeling compression
schemes (like min-peeling) are acyclic. So compression schemes arising
from sweeping across simple arrangements of hyperplanes in Euclidean or
Hyperbolic space are also acyclic. Does acyclicity characterize such compression schemes? \\

We have established peeling of all finite maximum and a family of infinite
maximum classes by representing them as PL hyperplane arrangements and
sweeping by multiple generic hyperplanes. A larger class of arrangements has
these properties---namely those which are $d$-contractible---and we have shown
that the corresponding one-inclusion graphs are precisely the strongly
contractible ones.  Finally we have established that there are $d$-maximal
classes that cannot be embedded in any $(d+k)$-maximum classes for any constant
$k$. Some important open problems along these lines are the following.

\begin{question}
Prove peeling of maximum classes using purely combinatorial arguments
\end{question}

\begin{question}
Can all maximal classes be peeled by representing them by hyperplane
arrangements and then using a sweeping technique (potentially solving the
Sample Compressibility conjecture)? The obvious candidate for this approach is to use
$d$-contractible PL hyperplane arrangements. 
\end{question}

\begin{question}
What about more general collections of infinite maximum classes, or
infinite arrangements?
\end{question}

\begin{question}
Is it true that any $d$-contractible PL hyperplane arrangement is
equivalent to a Hyperbolic hyperplane arrangement?
\end{question}

\begin{question}
Is it true that all strongly contractible classes, with largest
dimension $d$ of cubes can be embedded in maximum classes of VC dimension $d$? 
\end{question}

\acks{We thank Peter Bartlett for his very helpful feedback, and gratefully
acknowledge the support of the NSF through grants DMS-0434383 and DMS-0707060,
and the support of the Siebel Scholars Foundation.}

\vskip 0.2in
\bibliography{RR-compress-jmlr09-arxiv}

\end{document}